\documentclass{article}

\usepackage[final,nopatch=footnote]{microtype}
\usepackage{graphicx}
\usepackage{subcaption}
\usepackage{booktabs} 
\usepackage{caption}
\usepackage{multirow}

\usepackage{mathtools}

\usepackage{xcolor}
\definecolor{mycitecolor}{HTML}{3498DC}
\definecolor{mylinkcolor}{HTML}{E74D3B}
\definecolor{myurlcolor}{HTML}{980000}
\definecolor{mydarkgreen}{HTML}{6a994e}
\definecolor{myorange}{HTML}{E7730D}
\definecolor{myblue}{HTML}{4594c1}
\usepackage{soul}  
\newcommand{\hlc}[2][yellow]{\sethlcolor{#1}\hl{#2}}

\usepackage[most]{tcolorbox}
\newcommand{\Appendix}{\textcolor{mylinkcolor}{Appendix}}
\newcommand{\Figure}{\textcolor{mylinkcolor}{Figure}}
\newcommand{\Section}{\textcolor{mylinkcolor}{Section}}
\newcommand{\DAWN}{\textbf{\texttt{DAWN}}}
\newcommand{\Table}{\textcolor{mylinkcolor}{Table}}
\newcommand{\Equation}{\textcolor{mylinkcolor}{Equation}}
\usepackage[normalem]{ulem}
\usepackage{enumitem}
\usepackage{bm}

\newcommand{\Takeaway}[1]{%
\begin{tcolorbox}[
    enhanced,
    colback=white,
    colframe=white,
    leftrule=0.4mm,
    rightrule=0.4mm,
    toprule=0.4mm,
    bottomrule=0.4mm,
    arc=0mm,
    left=0pt,
    right=0pt,
    top=2pt,
    bottom=2pt,
    breakable,
    borderline north={0.4mm}{0pt}{mydarkgreen!80!black},
    borderline south={0.4mm}{0pt}{mydarkgreen!80!black}
]
{\textbf{\textcolor{mydarkgreen!80!black}{Takeaway:}}} #1
\end{tcolorbox}
}

\definecolor{dawnorange}{HTML}{E8862F}

\newcommand{\dawnbox}[1]{%
\begin{tcolorbox}[
    enhanced,
    colback=dawnorange!3!white,
    colframe=dawnorange,
    leftrule=2mm,
    rightrule=0mm,
    toprule=0mm,
    bottomrule=0mm,
    arc=0mm,
    left=5pt,
    right=5pt,
    top=5pt,
    bottom=5pt,
    breakable,
    leftlower=2mm,
    leftupper=2mm
]
\normalsize 
#1
\end{tcolorbox}
}

\newcommand{\contribution}[1]{%
\begin{tcolorbox}[
    enhanced,
    colback=myblue!8!white,
    colframe=myblue,
    leftrule=2mm,
    rightrule=0mm,
    toprule=0mm,
    bottomrule=0mm,
    arc=0mm,
    left=5pt,
    right=5pt,
    top=5pt,
    bottom=5pt,
    breakable,
    leftlower=2mm,
    leftupper=2mm
]
\normalsize 
#1
\end{tcolorbox}
}

\usepackage[preprint]{icml2026}

\usepackage{algorithm}

\usepackage[colorlinks=true,linkcolor=mylinkcolor,citecolor=mycitecolor,urlcolor=myurlcolor]{hyperref}
\hypersetup{hypertexnames=false}

\usepackage{titletoc}

\usepackage{amsmath}
\usepackage{amssymb}
\usepackage{mathtools}
\usepackage{amsthm}
\usepackage{pifont}

\usepackage[capitalize,noabbrev]{cleveref}

\theoremstyle{plain}

\theoremstyle{definition}

\theoremstyle{remark}

\usepackage[textsize=tiny]{todonotes}

\icmltitlerunning{What Makes Value Learning Efficient in Residual RL?}

\begin{document}

\twocolumn[
  \icmltitle{What Makes Value Learning Efficient in Residual Reinforcement Learning?}




  \icmlsetsymbol{equal}{*}

\begin{icmlauthorlist}
\icmlauthor{Guozheng Ma}{ntu}
\icmlauthor{Lu Li}{mila,udem}
\icmlauthor{Haoyu Wang}{ntu}
\icmlauthor{Zixuan Liu}{nus}
\icmlauthor{Pierre-Luc Bacon}{mila,udem}
\icmlauthor{Dacheng Tao}{ntu}
\end{icmlauthorlist}

\icmlaffiliation{ntu}{Nanyang Technical University}
\icmlaffiliation{mila}{Mila - Quebec AI Institute}
\icmlaffiliation{udem}{Université de Montréal}
\icmlaffiliation{nus}{National University of Singapore}

\icmlcorrespondingauthor{Guozheng Ma}{GUOZHENG001@e.ntu.edu.sg}
\icmlcorrespondingauthor{Dacheng Tao}{dacheng.tao@ntu.edu.sg}


  \vskip 0.3in
]



\printAffiliationsAndNotice{}  

\begin{abstract}
Residual reinforcement learning (RL) enables stable online refinement of expressive pretrained policies by freezing the base and learning only bounded corrections.
However, value learning in residual RL poses unique challenges that remain poorly understood.
In this work, we identify two key bottlenecks: \textit{cold start pathology}, where the critic lacks knowledge of the value landscape around the base policy, and \textit{structural scale mismatch}, where the residual contribution is dwarfed by the base action.
Through systematic investigation, we uncover the mechanisms underlying these bottlenecks, revealing that simple yet principled solutions suffice: base-policy transitions serve as an essential value anchor for implicit warmup, and critic normalization effectively restores representation sensitivity for discerning value differences.
Based on these insights, we propose \DAWN~(Data-Anchored Warmup and Normalization), a minimal approach targeting efficient value learning in residual RL.
By addressing these bottlenecks, \DAWN~demonstrates substantial efficiency gains across diverse benchmarks, policy architectures, and observation modalities.
\end{abstract}

\section{Introduction}

\begin{figure}[t]
\centering
\vspace{-0.5\baselineskip}
\begin{center}
\includegraphics[width=\linewidth]{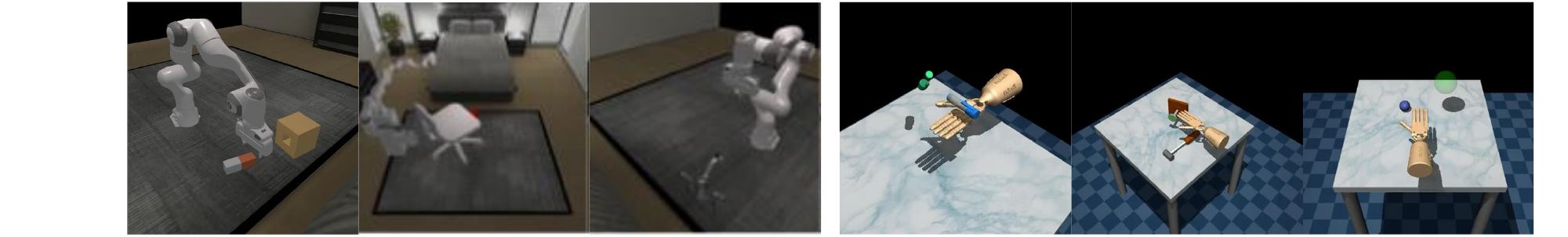}\\
\vspace{-0.2\baselineskip}
\includegraphics[width=\linewidth]{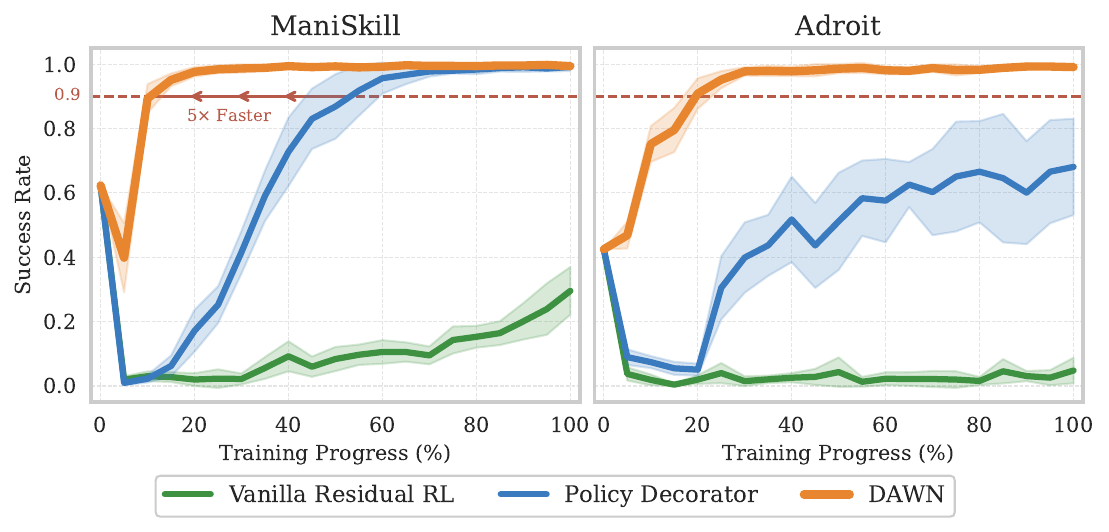}
\end{center}
\vspace{-1.1\baselineskip}
\caption{\textbf{\DAWN~enables efficient value learning in residual RL.} Aggregated success rates with Diffusion Policy as base policy across ManiSkill (3 tasks) and Adroit (3 tasks) benchmarks. \DAWN~achieves comparable final performance while converging approximately 5× faster than prior methods.}
\vspace{-1.5\baselineskip}
\label{fig:overall_aggregated}
\end{figure}

Generative policies pretrained on large-scale demonstrations, from Diffusion Policy~\citep{chi2025diffusion, song2025survey} to Vision-Language-Action (VLA) models~\citep{black2024pi_0, intelligence2025pi}, have reshaped the landscape of robot learning.
Despite their expressiveness, imitation alone cannot meet the demands of reliable deployment due to inevitable distribution shift and compounding errors~\citep{pan2026sop, li2025vla}.
Simply scaling demonstration data cannot resolve this dilemma, since only interaction equips policies with the ability to recover through trial-and-error~\citep{silver2025welcome}. 
This makes \textbf{online reinforcement learning (RL)} essential for last-mile reliability, particularly in precision-critical domains such as contact-rich manipulation~\citep{lei2025rl}.


Among available approaches, residual RL offers a principled path for online refinement~\citep{silver2018residual, johannink2019residual}: freezing the base policy and learning only a bounded correction avoids the forgetting and instability of modifying pretrained parameters~\citep{jiang2024transic, zhou2025efficient}.
Recent work has advanced the paradigm through policy-centric innovations, from foundational techniques like bounded residuals~\citep{ankile2025imitation} and progressive exploration~\citep{yuan2025policy}, to finer-grained designs such as per-step corrections~\citep{ankile2025residual} and latent steering for mode selection in multimodal policies~\citep{anonymous2026unified}.
These efforts address the central question of \textbf{\textit{how to design an effective and stable residual policy}}, making the paradigm practical for real-world refinement~\citep{wang2025residualmppi, xiao2025self}.


However, effective policy design alone does not guarantee sample efficiency.
Even state-of-the-art methods like \citet{yuan2025policy} require over a million online interactions, a sample complexity prohibitive for real-world deployment.
We argue that this gap stems from an overlooked factor: \textbf{\textit{value learning}}.
In the off-policy actor-critic methods underlying residual RL, policy improvement depends entirely on critic gradients~\citep{haarnoja2018soft}.
If value learning is inefficient, no amount of policy engineering can compensate. 
This motivates three core research questions:
\vspace{-0.2\baselineskip}
\contribution{
\begin{enumerate}[leftmargin=2.5em, itemsep=0em]
    \item[\textbf{\textit{\textcolor{myblue}{RQ1:}}}]
    What are the unique and critical challenges for value learning in residual RL?
    \item[\textbf{\textit{\textcolor{myblue}{RQ2:}}}]
    What bottlenecks underlie these challenges, and how can they be fundamentally resolved?
    \item[\textbf{\textit{\textcolor{myblue}{RQ3:}}}]
    What gains in sample efficiency can be achieved through principled solutions?
\end{enumerate}
}
\vspace{-0.4\baselineskip}


Our investigation reveals two challenges unique to value learning in residual RL, each addressed by a simple but principled solution.
\textcolor{mydarkgreen}{$\bullet$}~\textbf{\textit{Cold Start Pathology}}: The critic begins without knowledge of the value landscape around the base policy. We find that transitions from the base policy serve as a necessary and sufficient \textit{value anchor}, enabling effective implicit warmup. Explicit critic pre-training, by contrast, fails due to pathological dynamics arising from the entropy mechanism (\Section~\ref{sec:cold_start}).
\textcolor{mydarkgreen}{$\bullet$}~\textbf{\textit{Structural Scale Mismatch}}: The residual contribution is dwarfed by the base action, making it difficult for the critic to discern value differences. Normalization addresses this by restoring \textit{critic sensitivity}, enabling effective credit assignment to residual actions. Despite input-level suppression, the residual's effect on Q-values manifests as a clear mean shift, making distributional objectives unnecessary (\Section~\ref{sec:scale_mismatch}).
By understanding these bottlenecks, the above insights naturally translate into our method, \DAWN~(Data-Anchored Warmup and Normalization). 
Focusing on value learning in residual RL, \DAWN~substantially improves sample efficiency, as illustrated in \Figure~\ref{fig:overall_aggregated}.
In summary, our contributions are:

\vspace{-0.2\baselineskip}
\contribution{
\begin{enumerate}[leftmargin=*, itemsep=0.2em]
\item We provide the first investigation of value learning in residual RL, identifying two unique challenges: \textit{cold start pathology} and \textit{structural scale mismatch}.
    
\item We uncover the mechanisms underlying these challenges through extensive analysis, revealing that warmup data provides an essential value anchor and normalization effectively restores critic sensitivity.
    
\item We propose \DAWN, a principled yet minimal approach that is easy to implement. 
Extensive experiments across diverse benchmarks, policy architectures, and observation modalities demonstrate substantial improvements in sample efficiency.
\end{enumerate}
}
\vspace{-0.5\baselineskip}
\section{Preliminary: Residual RL}
\label{sec:preliminary}

Residual RL is built on a simple idea: instead of full policy optimization, learn only a small correction on top of an already capable base policy. 
This reduces the problem from policy search to local refinement around a reliable prior. The paradigm originated from augmenting hand-crafted controllers with learned corrections~\citep{silver2018residual, johannink2019residual}, and has since evolved to handle expressive policies trained through imitation. 
With the emergence of expressive policy models such as Diffusion Policy~\citep{chi2025diffusion} and Vision-Language-Action models~\citep{kim2024openvla, black2024pi_0}, residual RL becomes particularly appealing for precision-critical tasks such as robotic manipulation. 
It enables stable online refinement of powerful offline-trained policies while avoiding the prohibitive sample complexity of learning from scratch.

\textbf{Problem Formulation.}~
In residual RL, the executed action combines a frozen base policy with a learned residual:
\begin{equation}
a = a_\text{base}+\lambda \cdot a_\text{res} = \underbrace{\pi_{\text{base}}(s)}_{\mathclap{\substack{\text{Pre-Trained} \\ \text{Frozen}}}} +  \lambda \cdot \underbrace{\pi_{\text{res}}(s)}_{\mathclap{\substack{\text{Randomly Initialized} \\ \text{Learnable}}}}
\label{eq:residual_action}
\end{equation}
where $\pi_{\text{base}}$ is typically obtained through imitation learning and remains fixed during RL training, while $\pi_{\text{res}}$ is a lightweight network (e.g., a small MLP) that learns task-specific corrections. The scaling factor $\lambda$ bounds the residual's influence and is typically set to a small value such as $0.1$.
The critic takes the summed action as input, i.e., $Q(s, \pi_{\text{base}}(s) + \lambda \cdot \pi_{\text{res}}(s))$. This formulation, as opposed to concatenating base and residual actions separately, has been empirically validated as the most effective choice~\citep{yuan2025policy, ankile2025residual}. 
However, the small $\lambda$ means that the residual's contribution is dwarfed by the base action, creating a \textit{structural scale mismatch} that makes it difficult for the critic to distinguish value differences caused by $\pi_{\text{res}}$, a challenge we analyze in \Section~\ref{sec:scale_mismatch}.

A second challenge is \textit{value cold start}. As an imitation-to-online method, residual RL must learn the value function from scratch since no offline data is available for critic pre-training. What makes residual RL distinct is that the base policy remains frozen, and all adaptation must proceed through a randomly initialized residual $\pi_{\text{res}}$. We examine how this structure affects value learning in \Section~\ref{sec:cold_start}.

\textbf{A Minimal Baseline for Investigation.}~
To isolate the core challenges of value learning, we adopt a minimal residual RL setup using SAC~\citep{haarnoja2018soft} as the underlying off-policy actor-critic algorithm, which incorporates entropy regularization to encourage exploration. 
The residual scale $\lambda$ in \Equation~\ref{eq:residual_action} is the only task-specific hyperparameter, for which we follow the values from~\citet{yuan2025policy}.
We deliberately avoid additional stabilization mechanisms to expose the raw dynamics of value learning. Full implementation details are provided in \Appendix~\ref{app:baseline}.
\begin{figure*}[ht]
\centering
\begin{center}
\vspace{-0.5\baselineskip}
\includegraphics[width=\linewidth]{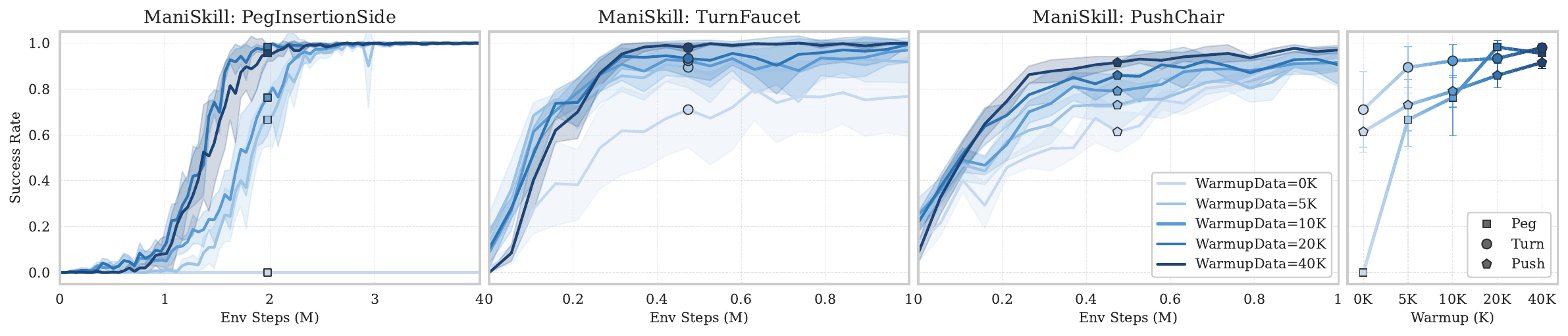}
\end{center}
\vspace{-0.9\baselineskip}
\caption{\textbf{Effect of warmup data quantity on learning performance.} (Left three) Learning curves across three ManiSkill tasks with varying amounts of warmup data. (Right) Success rate at the midpoint of training versus warmup data quantity. More warmup data consistently improves sample efficiency, with the effect most pronounced on challenging tasks. All experiments use 8 random seeds with shaded regions indicating 95\% confidence intervals, a convention we follow throughout the paper.}
\vspace{-1.3\baselineskip}
\label{fig:warmup_data}
\end{figure*}

\vspace{-0.2\baselineskip}
\section{Dissecting Value Learning in Residual RL}
\label{sec:dissecting}
\vspace{-0.2\baselineskip}

In residual RL, policy improvement relies entirely on critic gradients, making accurate value learning the foundation of efficient adaptation. 
To unlock this potential, this section dissects two challenges unique to value learning in residual RL: the cold start pathology (\Section~\ref{sec:cold_start}) and the structural scale mismatch (\Section~\ref{sec:scale_mismatch}). 
For each challenge, we examine two candidate solutions motivated by different intuitions and analyze their effectiveness. 
Our investigation demonstrates that effectively addressing these challenges substantially improves sample efficiency, while providing insights into the underlying mechanisms and the principles behind effective solutions.
All experiments use the minimal baseline described in \Section~\ref{sec:preliminary} on ManiSkill~\citep{gu2023maniskill2} tasks with Diffusion Policy~\citep{chi2025diffusion} as the base policy; full details are provided in \Appendix.

\vspace{-0.2\baselineskip}
\subsection{Challenge I: The Cold Start Pathology}
\label{sec:cold_start}
\vspace{-0.2\baselineskip}

At the start of training, the critic is randomly initialized and has no knowledge of the value landscape around the base policy.  
In online RL from scratch, value estimates and policy co-evolve from random initialization, allowing the critic to mature alongside policy improvements.  
In residual RL, however, the full policy operates near a competent base policy from the outset.
If the critic cannot quickly ground itself to this region, erroneous value estimates will misguide the residual policy, risking catastrophic performance collapse.

Two intuitions suggest potential remedies. 
\textcolor{mydarkgreen}{$\bullet$}~The first perspective emphasizes \textit{data as implicit warmup}: effective value learning may require sufficient warmup transitions before training begins.
\textcolor{mydarkgreen}{$\bullet$}~The second emphasizes \textit{explicit warmup training}: perhaps the critic requires dedicated pre-training before the residual policy starts updating.
We investigate both perspectives below.


\vspace{-0.2\baselineskip}
\subsubsection{The necessity of warmup data as anchor}
\label{sec:warmup_anchor}

We begin by examining how the amount of warmup data affects learning. Before online training, we collect transitions using the base policy and store them in the replay buffer. The critic then learns from this data alongside newly collected transitions once training starts.

\begin{figure}[t]
\centering
\vspace{-0.2\baselineskip}
\begin{center}
\includegraphics[width=\linewidth]{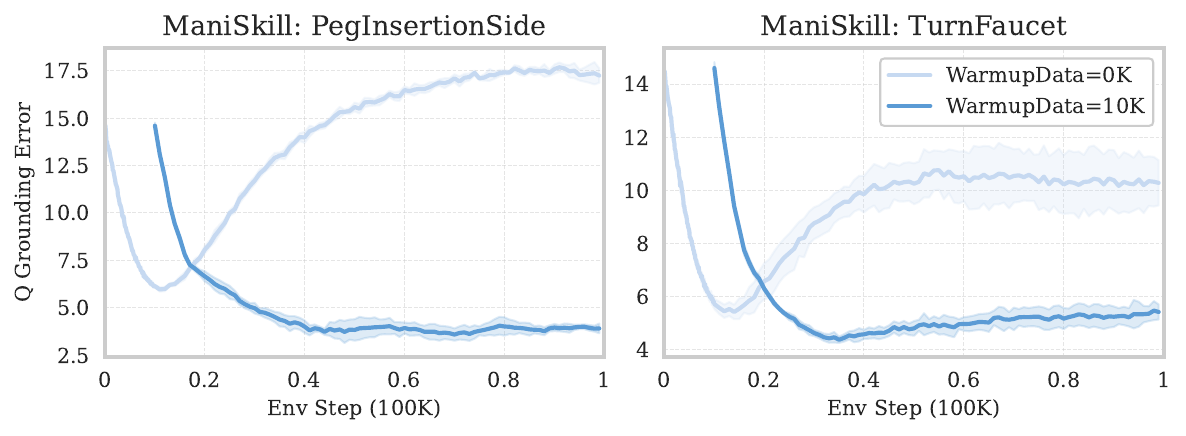}
\end{center}
\vspace{-1\baselineskip}
\caption{\textbf{Q-value grounding error during early training.} Without 
warmup data, the error briefly decreases but quickly diverges. With warmup, 
the critic maintains accurate estimates throughout, confirming the value 
anchor effect.}
\vspace{-1.7\baselineskip}
\label{fig:grounding}
\end{figure}

\begin{figure*}[t]
\centering
\vspace{-0.5\baselineskip}
\begin{center}
\includegraphics[width=\linewidth]{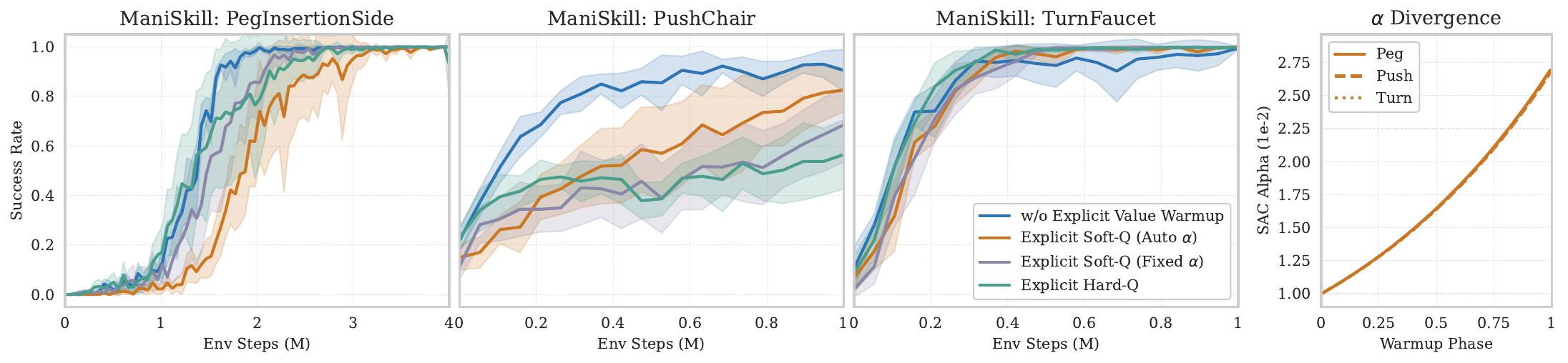}
\end{center}
\vspace{-\baselineskip}
\caption{\textbf{Effect of explicit value warmup on learning performance.} (Left three) Explicit warmup variants fail to improve and often degrade sample efficiency compared to implicit warmup alone. (Right) With automatic entropy tuning, $\alpha$ diverges during the warmup phase across all tasks, even with an initial value as small as $0.01$. Larger initial values lead to more severe divergence (see \Appendix).
}
\vspace{-1.5\baselineskip}
\label{fig:explicit_warmup}
\end{figure*}

\textbf{Effect on Learning Performance.~}
\Figure~\ref{fig:warmup_data} shows that warmup data substantially affects learning efficiency. 
Without warmup, the most challenging task (PegInsertionSide) fails entirely, while easier tasks suffer significant sample inefficiency. 
Performance improves consistently as more warmup transitions are provided. Based on these results, we use 20K warmup transitions as default in subsequent experiments.
This finding aligns with concurrent work~\citep{ankile2025residual}, which shows that incorporating offline demonstrations into the replay buffer improves stability in residual RL. Our result extends this insight: rather than requiring pre-collected demonstrations with reward labels, simply collecting transitions from the base policy is sufficient.

\textbf{The Value Anchor Effect.}~
Seeding the replay buffer with initial transitions is standard practice in off-policy RL~\citep{lillicrap2015continuous}. 
However, with a random policy, such data carries limited information and is typically kept minimal. 
In residual RL, the situation differs: warmup data collected from the base policy provides meaningful trajectories as a prior for value learning. 
Specifically, since residual RL refines the base policy through local corrections rather than global policy search, the critic must accurately capture the value landscape in the neighborhood of the base policy.
Warmup data serves as a \textit{value anchor} that grounds the critic to this region before residual learning begins.

To verify this anchor effect, we measure the \textit{Q-value grounding error}: 
the discrepancy between the critic's estimates and the true returns on base 
policy trajectories. 
Concretely, we compute $\mathcal{E}_{\text{grounding}} = |Q_\theta(s, a_{\text{base}}) - G^{\pi_{\text{base}}}|$,
where $G^{\pi_{\text{base}}}$ is the Monte Carlo return. 
This metric directly measures how well the critic captures the value landscape in the region where residual learning operates. Details on the experimental design and formal justification are in \Appendix~\ref{app:grounding_error}.
\Figure~\ref{fig:grounding} reveals a clear contrast:
\textcolor{mydarkgreen}{$\bullet$}~Without warmup data, the grounding error briefly decreases but quickly diverges, indicating the critic fails to establish stable value estimates.
\textcolor{mydarkgreen}{$\bullet$}~With warmup data, the error drops rapidly and remains low, confirming that warmup data successfully anchors the critic to the relevant value landscape.

A natural question is whether adding exploration during warmup would improve coverage and accelerate learning. We compare the base policy alone against several exploration strategies (details in \Appendix~\ref{app:warmup_strategy}).
Surprisingly, the base policy alone matches or exceeds all alternatives. 
This reinforces the value anchor interpretation: grounding the critic matters more than collecting diverse trajectories.
Exploration emerges naturally once training begins through the stochastic policy and entropy regularization.

\vspace{-0.3\baselineskip}
\Takeaway{
Warmup data addresses cold start by serving as a \textbf{value anchor}: providing a prior over the value landscape around the base policy is both necessary and sufficient for bootstrapping efficient value learning.
}
\vspace{-0.7\baselineskip}

\subsubsection{The redundancy of explicit warmup phase}
\label{sec:explicit_warmup}

Given that warmup data provides an effective value anchor, a natural question arises: \textit{\textbf{can explicit critic pre-training further accelerate learning?}} 
We investigate this by freezing the residual policy and training only the critic during an initial warmup phase. 
We consider three representative variants that span the main design choices:
\textcolor{mydarkgreen}{$\bullet$}~\textit{Explicit Soft Q (Auto $\alpha$)}: standard SAC critic updates with automatic entropy tuning;
\textcolor{mydarkgreen}{$\bullet$}~\textit{Explicit Soft Q (Fixed $\alpha$)}: critic updates with entropy coefficient fixed at its initial value; and
\textcolor{mydarkgreen}{$\bullet$}~\textit{Explicit Hard Q}: critic updates using the standard Bellman target without entropy regularization.
Contrary to expectations, \Figure~\ref{fig:explicit_warmup} shows that all explicit warmup variants fail to improve sample efficiency and often make it worse. We analyze the failure modes of each approach below.

\begin{figure}[t]
\centering
\includegraphics[width=\linewidth]{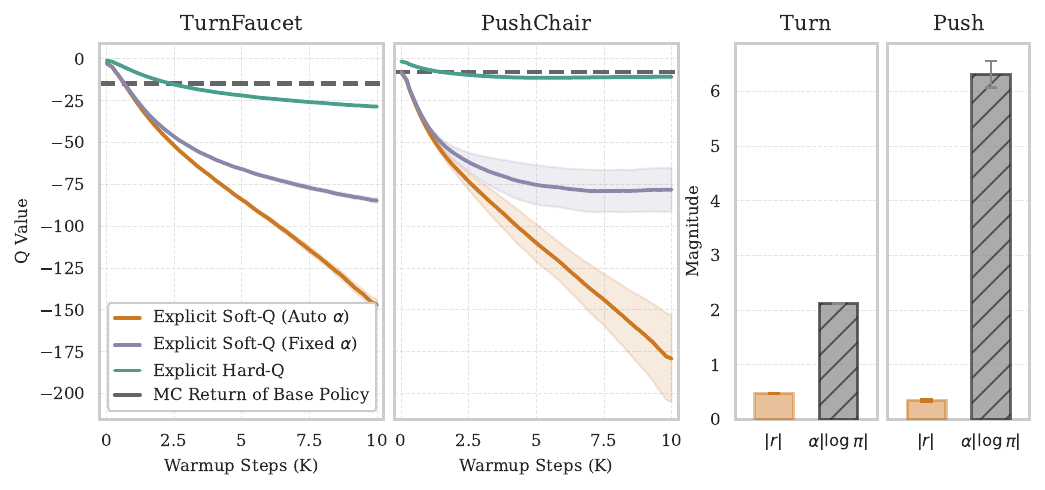}
\vspace{-1.5\baselineskip}
\caption{\textbf{The failure mechanism of explicit value warmup.} (Left) Q-value estimates during the warmup phase. Soft Q methods collapse to extreme negative values, while Hard Q remains near the true MC return. (Right) During explicit Soft Q warmup, the magnitude of $|\alpha \log \pi|$ substantially exceeds $|r|$.}
\label{fig:entropy_dominate}
\vspace{-1.7\baselineskip}
\end{figure}

\textbf{Entropy Dominance and Value Collapse.}~
With autotuned $\alpha$, the entropy temperature diverges during warmup (\Figure~\ref{fig:explicit_warmup}), a phenomenon also observed in~\citet{yuan2025policy}. However, even with fixed $\alpha$, explicit Soft Q warmup suffers from a more fundamental problem unique to residual RL.

As shown in \Figure~\ref{fig:entropy_dominate} (left), the Soft Q values collapse to extreme negative values during the warmup phase, far below the true Monte Carlo returns. 
This collapse stems from \textit{entropy dominance}: when $|\alpha \log \pi|$ substantially exceeds $|r|$ (\Figure~\ref{fig:entropy_dominate}, right), the TD target is dominated by the entropy term rather than the task reward. 
Consequently, this large entropy term introduces a strong negative bias in the Q target, causing the Q-value network to fit this bias rather than the true value landscape.
This is precisely why Soft Q warmup exhibits severe value collapse.

This pathology arises from the initialization of residual RL. To avoid disrupting the base policy at the start of training, the residual policy $\pi_{\text{res}}$ is initialized to output near-zero actions with minimal variance, making it nearly deterministic. 
This makes $|\log \pi|$ extremely large, so even a small $\alpha$ yields an entropy term that dwarfs the task reward by an order of magnitude, defeating the purpose of explicit warmup.

\textbf{Why Hard Q Warmup Is Not the Solution.}~
Given the entropy-induced failure of Soft Q warmup, a natural alternative is to remove entropy regularization entirely.
However, while Hard Q avoids value collapse, it introduces an \textit{objective mismatch}: the critic is pre-trained with Hard Q targets but subsequent training uses Soft Q learning. This inconsistency prevents Hard Q from providing meaningful benefits.

\vspace{-0.4\baselineskip}
\Takeaway{
Explicit value warmup is \textit{\textbf{redundant}} in residual RL. The near-deterministic initialization of $\pi_{\text{res}}$ causes entropy dominance that corrupts value learning, while removing entropy creates objective mismatch. 
The implicit warmup through data anchoring alone is \textbf{\textit{both simpler and more effective}} for addressing the cold start pathology.
}

\subsection{Challenge II: The Structural Scale Mismatch}
\vspace{-0.5em}
\label{sec:scale_mismatch}

The bounded residual formulation (\Equation~\ref{eq:residual_action}) uses a small $\lambda$ to ensure that residual corrections remain local~\citep{yuan2025policy}. 
However, this introduces a structural challenge for value learning that is unique to residual RL.

Unlike standard RL where the entire action is learnable, residual RL introduces a structural asymmetry: the frozen base action dominates in magnitude, while the learnable residual is suppressed by a factor of $\lambda$.
When this asymmetric combination $a = a_{\text{base}} + \lambda \cdot a_{\text{res}}$ serves as input to the critic, the resulting \textit{scale mismatch} poses a fundamental challenge for credit assignment. To guide policy improvement, the critic must distinguish value differences caused by different residual actions.
However, $a_{\text{base}}$ dominates the input magnitude, making it difficult for the critic to correctly attribute value to the residual action.

To address this scale mismatch, we investigate two methodological paths. \textcolor{mydarkgreen}{$\bullet$}~The first focuses on \textit{architectural intervention}: employing normalization techniques to decouple the critic's internal representations from input magnitudes, allowing it to remain sensitive to small variations regardless of the dominant base action. 
\textcolor{mydarkgreen}{$\bullet$}~The second considers \textit{objective refinement}: distributional RL provides richer learning signals by modeling the full return distribution, potentially capturing value distinctions that mean-based critics miss.

\vspace{-0.3em}
\subsubsection{Restoring Sensitivity via Normalization}
\vspace{-0.3em}
\label{sec:normalization}
Given the scale mismatch identified above, we explore whether normalization techniques can restore the critic's sensitivity to residual actions. We examine two representative approaches: Layer Normalization (LN)~\citep{ba2016layer} and Hyperspherical Normalization (HN)~\citep{lee2025hyperspherical}.

LN normalizes activations across features within each layer:
\vspace{-0.2\baselineskip}
\begin{equation}
    \text{LN}(\mathbf{x}) = \gamma \odot \frac{\mathbf{x} - \mu}{\sigma} + \beta
\end{equation}
where $\mu$ and $\sigma$ are the mean and standard deviation computed over the feature dimension, and $\gamma$ and $\beta$ are learnable affine parameters. This stabilizes the activation statistics regardless of input magnitude.

Unlike LN which normalizes statistical moments~\citep{lee2025simba}, HN~\citep{lee2025hyperspherical} enforces a strict geometric constraint. 
For a layer with input $\mathbf{x} \in \mathbb{R}^d$ and weight $\mathbf{w} \in \mathbb{R}^d$, both are projected onto the unit hypersphere:
\begin{equation}
    \hat{\mathbf{x}} = \mathbf{x} / \|\mathbf{x}\|_2, \quad 
    \hat{\mathbf{w}} = \mathbf{w} / \|\mathbf{w}\|_2
    \label{eq:hn_projection}
\end{equation}
The pre-activation output then depends solely on the angle $\theta$ between vectors, independent of their original magnitudes:
\begin{equation}
    y = s \cdot (\hat{\mathbf{w}}^T \hat{\mathbf{x}}) = s \cdot \cos(\theta)
    \label{eq:hn_output}
\end{equation}
where $s$ is a learnable scalar. By explicitly removing magnitude information, 
HN ensures the critic responds only to directional changes in the input.

\textbf{Effect on Learning Efficiency.}~ 
Building on the warmup data foundation, we evaluate whether normalization can address the scale mismatch challenge. 
\begin{figure}[ht]
\centering
\vspace{-\baselineskip}
\begin{center}
\includegraphics[width=\linewidth]{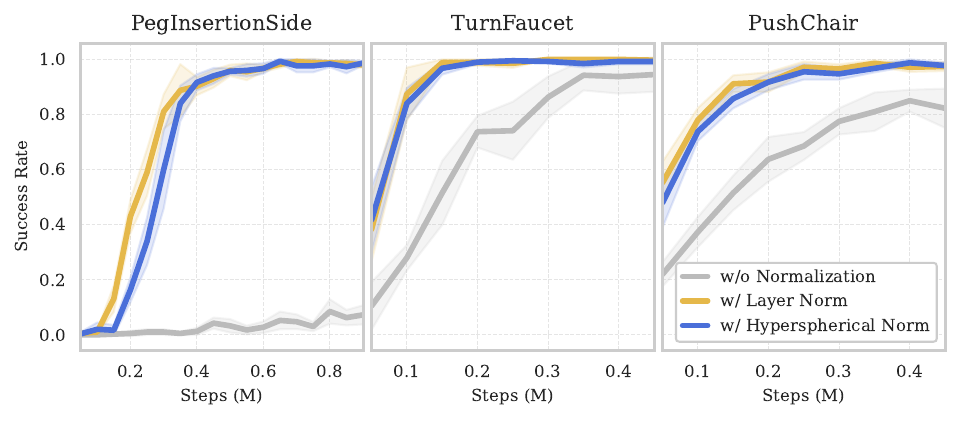}
\end{center}
\vspace{-1.2\baselineskip}
\caption{\textbf{Effect of critic normalization.} Both LN and HN substantially improve learning efficiency compared to the unnormalized baseline, with comparable performance to each other.}
\label{fig:norm_effect}
\vspace{-1.5\baselineskip}
\end{figure}

\Figure~\ref{fig:norm_effect} compares sample efficiency with and without normalization applied to the critic network. Both LN and HN substantially improve sample efficiency across all tasks, with the most pronounced gains on the PegInsertionSide task. Notably, although LN and HN operate through different mechanisms (statistical normalization versus geometric projection), they achieve comparable performance. This suggests that the key factor is restoring sensitivity to residual variations, not the specific normalization strategy.





\begin{figure}[ht]
\centering
\vspace{-\baselineskip}
\begin{center}
\includegraphics[width=\linewidth]{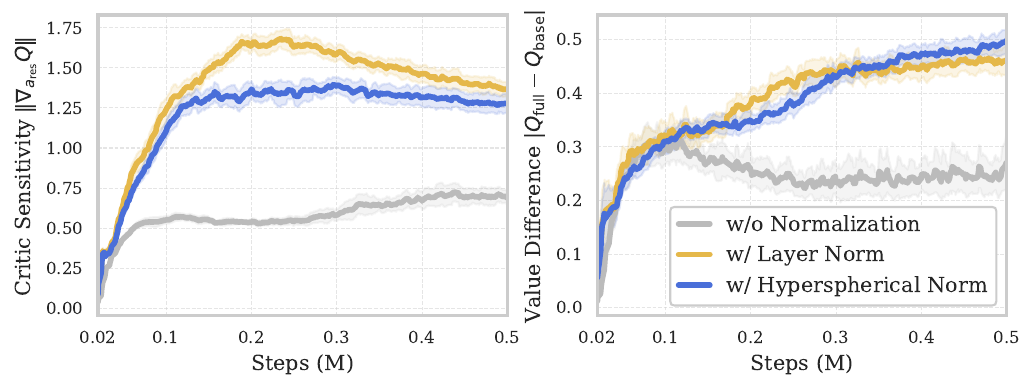}
\end{center}
\vspace{-1\baselineskip}
\caption{\textbf{Mechanism analysis of normalization.} \textbf{\textit{(Left)}} Critic sensitivity to residual actions, measured by $\|\nabla_{a_{\text{res}}} Q\|$. Normalization yields significantly higher sensitivity, enabling the critic to detect small residual variations. \textbf{\textit{(Right)}} Value contribution of the residual policy, measured by $|Q_{\text{full}} - Q_{\text{base}}|$. With normalization, the critic learns to attribute increasing value to the residual throughout training. \textit{Evaluated on PegInsertionSide.}} 
\vspace{-1\baselineskip}
\label{fig:mechanism}
\end{figure}

\textbf{Mechanism: How Normalization Helps.}~
To understand why normalization improves learning, we analyze two diagnostic metrics on the PegInsertionSide task (\Figure~\ref{fig:mechanism}).

The left panel shows the critic's sensitivity to residual actions, measured by $\|\nabla_{a_{\text{res}}} Q\|$. This gradient norm quantifies how strongly the critic's output responds to changes in the residual. With normalization, sensitivity rises rapidly and reaches approximately twice the level of the unnormalized baseline, confirming that normalization enables the critic to better detect residual variations despite their small magnitude.
The right panel shows the estimated value contribution of the residual policy, measured by $|Q(s, a_{\text{base}} + \lambda \cdot a_{\text{res}}) - Q(s, a_{\text{base}})|$. This quantity captures the value difference the critic assigns to the residual action. With normalization, this contribution grows steadily throughout training, indicating that the critic learns to recognize the residual's effect on value. Without normalization, growth is substantially slower, suggesting the critic struggles to distinguish the residual's contribution.

\begin{figure}[t]
\centering
\vspace{-0.5\baselineskip}
\begin{center}
\includegraphics[width=\linewidth]{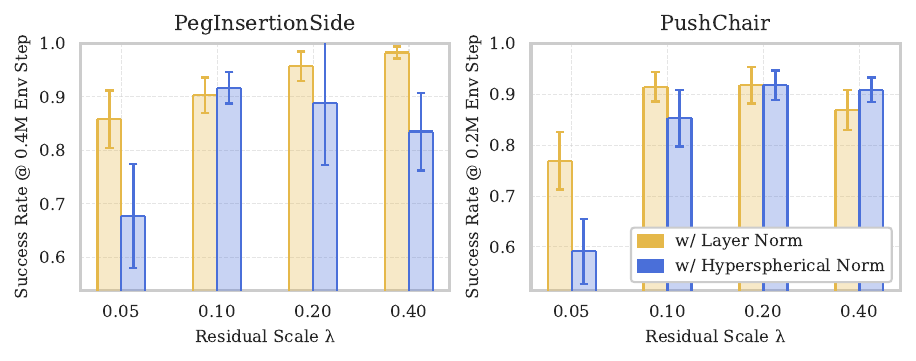}
\end{center}
\vspace{-0.8\baselineskip}
\caption{\textbf{Robustness to residual scale $\lambda$.} 
LN maintains stable performance across all $\lambda$ values, while HN exhibits higher variance and degrades significantly at small scales.} 
\label{fig:lambda_robust}
\vspace{-1.5\baselineskip}
\end{figure}

\textbf{Robustness to Residual Scale.}~
The residual scale $\lambda$ controls the magnitude of policy corrections and may vary across tasks. We compare LN and HN across a range of $\lambda$ values to assess their robustness.
As shown in \Figure~\ref{fig:lambda_robust}, while both methods perform well at moderate scales, HN degrades significantly at extreme values such as $\lambda = 0.05$.
In contrast, LN maintains stable performance across the entire range. 
Although HN has demonstrated strong performance in settings involving network scaling or massively parallel training~\citep{lee2025hyperspherical, seo2025fasttd3}, its strict normalization may be less suited to the fine-grained adaptation required in residual RL.
In this setting, LN's adaptive statistics provide greater flexibility. Combined with its simplicity, we adopt LN as the default for critic normalization.

\Takeaway{
Normalization effectively addresses the scale mismatch. Both statistical (LN) and geometric (HN) approaches restore critic sensitivity to residual actions, substantially improving value learning efficiency.
}

\subsubsection{Revisiting Distributional Objectives}
\label{sec:distributional}

Distributional RL models the full return distribution $Z(s,a)$ rather than just its expectation $Q(s,a) = \mathbb{E}[Z(s,a)]$, providing richer learning signals that have proven beneficial in large-scale and multi-task  scenarios~\citep{lee2025hyperspherical, nauman2025bigger}.
Given the scale mismatch challenge, where small residual variations must be distinguished in the value landscape, we investigate whether this additional expressiveness offers advantages over scalar regression.

We compare the standard MSE objective against three distributional alternatives, all implemented within the SAC framework by replacing the scalar critic with distributional variants:
\textcolor{mydarkgreen}{$\bullet$}~\textit{C51}~\citep{bellemare2017distributional}: models returns as a categorical distribution over a fixed set of atoms, trained with cross-entropy loss.
\textcolor{mydarkgreen}{$\bullet$}~\textit{QR}~\citep{dabney2018distributional}: learns a set of return quantiles via quantile regression, without requiring fixed support.
\textcolor{mydarkgreen}{$\bullet$}~\textit{TQC}~\citep{kuznetsov2020controlling}: extends quantile regression by truncating the highest estimates to mitigate overestimation.
While these methods differ in distributional representation, they share the goal of capturing richer information beyond the mean. 
Full implementation details are provided in \Appendix.

\begin{figure}[t]
\centering
\vspace{-0.6\baselineskip}
\begin{center}
\includegraphics[width=\linewidth]{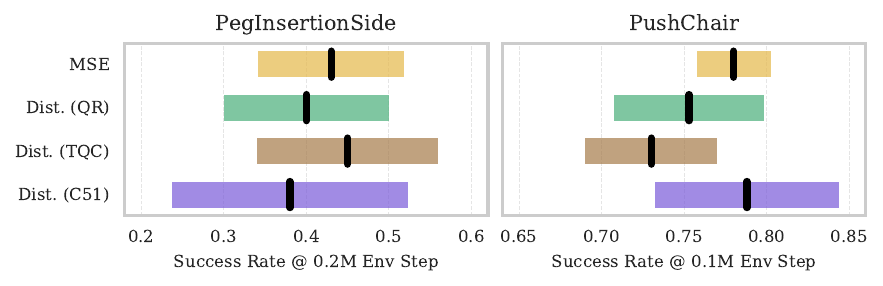}
\end{center}
\vspace{-1.1\baselineskip}
\caption{\textbf{Comparison of critic objectives.} MSE achieves competitive or superior early-stage performance compared to distributional methods (C51, QR, TQC), indicating no significant efficiency gains from modeling the full return distribution.}
\label{fig:objectives}
\vspace{-1.5\baselineskip}
\end{figure}

\textbf{Distributional Objectives Offer No Efficiency Gains.}
Building on the warmup and normalization foundations established above, all tasks converge to near-perfect success rates.
To assess whether distributional objectives provide additional benefits, we compare early-stage performance on two challenging tasks.
As shown in \Figure~\ref{fig:objectives}, MSE achieves competitive or superior efficiency compared to all distributional variants, indicating that the added complexity of modeling the full return distribution yields no significant improvement for residual value learning.

\begin{figure}[H]
\centering
\vspace{-0.5\baselineskip}
\begin{center}
\includegraphics[width=\linewidth]{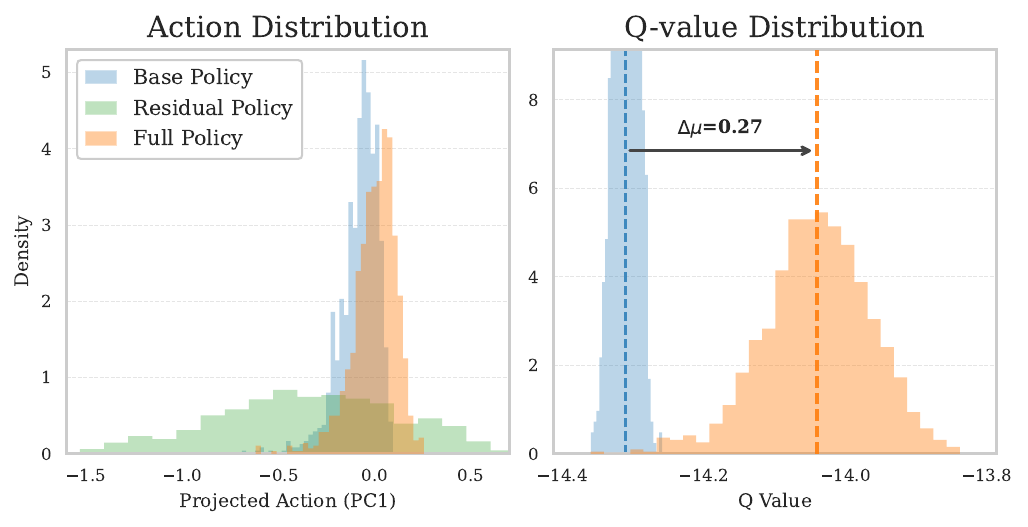}
\end{center}
\vspace{-1\baselineskip}
\caption{\textbf{The anatomy of residual value learning.} At a training checkpoint, we visualize action and Q-value distributions from 1024 samples. 
\textbf{\textit{(Left)}} Actions projected onto the first principal component. The residual induces a subtle shift, illustrating scale mismatch at the input level.
\textbf{\textit{(Right)}} Despite the small action shift, Q-value distributions show clear separation ($\Delta\mu = 0.27$), confirming that the residual's effect manifests as a distinct mean shift. \textit{Evaluated on PegInsertionSide with LN.}}
\label{fig:anatomy}
\vspace{-1.5\baselineskip}
\end{figure}

\textbf{Why MSE Suffices.}~
\Figure~\ref{fig:anatomy} provides mechanistic insight into why distributional objectives offer no advantage.
At a critical training checkpoint, we sample 1024 state-action pairs and visualize both the action distributions and the corresponding Q-value distributions produced by the critic.

\begin{figure*}[ht]
\centering
\vspace{-0.5\baselineskip}
\begin{center}
\includegraphics[width=\linewidth]{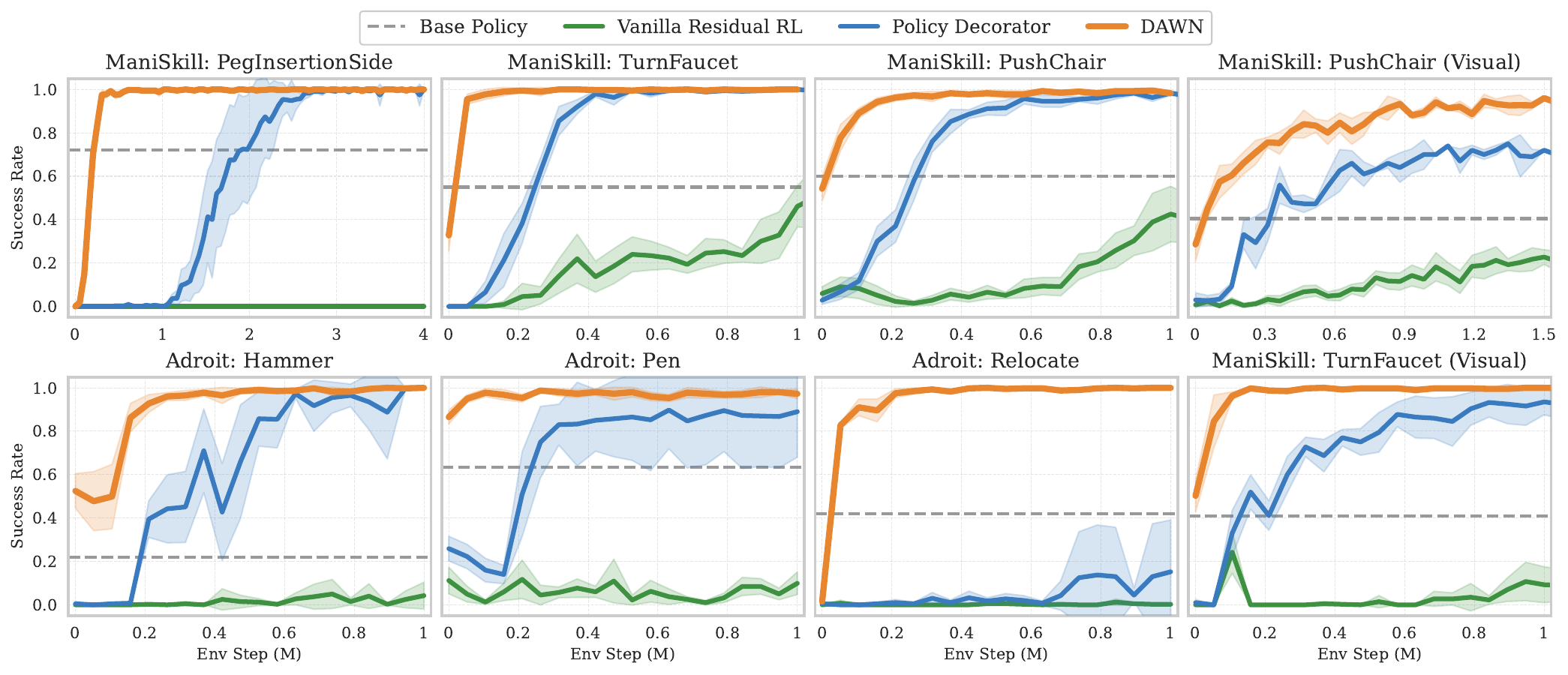}
\end{center}
\vspace{-0.8\baselineskip}
\caption{\textbf{Main results with Diffusion Policy as the base policy.} 
Across 8 tasks spanning ManiSkill and Adroit benchmarks (including 2 with visual observations), \DAWN~consistently achieves superior sample efficiency compared to all baselines, converging to high success rates significantly faster.
Dashed lines indicate base policy performance. 
Shaded regions denote 95\% confidence intervals over 8 seeds.}
\vspace{-1\baselineskip}
\label{fig:overall}
\end{figure*}

The left panel projects actions onto their first principal component. 
The residual introduces only a \textit{\textbf{marginal shift in the action distribution}}, leaving the full policy nearly indistinguishable from the base in action space.
This directly illustrates the scale mismatch challenge for value learning: the residual contribution is structurally suppressed, constituting only a bounded correction to the critic's action input.

The right panel shows the corresponding output: despite the marginal action shift, the \textit{\textbf{Q-value distributions exhibit clear separation}}.
Normalization enables the critic to translate subtle input differences into a distinct value signal. 
Importantly, this signal manifests as a pronounced \textit{mean shift} rather than a nuanced change in distribution shape.
Since MSE directly targets mean differences, it is well-suited to capture the value changes induced by residual policies.
In contrast, distributional methods must learn the entire distribution shape, a strictly harder objective that offers no efficiency advantage when the mean alone provides a sufficient learning signal.

\Takeaway{Despite the structurally suppressed residual contribution at the input level, its effect on Q-values manifests as a pronounced mean shift. This makes MSE the natural objective for residual value learning, with distributional methods offering no additional efficiency benefit.}
\section{\texttt{DAWN}: Unlocking Efficient Value Learning} 
\label{sec:dawn}

Efficient value learning in residual RL requires addressing two fundamental challenges: the \textit{cold start pathology} and the \textit{structural scale mismatch}.
We propose \DAWN~(\textbf{\texttt{D}}ata-\textbf{\texttt{A}}nchored \textbf{\texttt{W}}armup and \textbf{\texttt{N}}ormalization), which targets each challenge with a minimal, principled intervention:

\dawnbox{
\textbf{{Data-Anchored Warmup} $\Rightarrow$ \textit{Cold Start Pathology}}

\vspace{0.2em}

{\leftskip=1em
Before training, collect $M$ transitions using the base policy alone and store them in the replay buffer.\par}

\vspace{0.5em}

\textbf{{Critic Normalization} $\Rightarrow$ \textit{Structural Scale Mismatch}}
\vspace{0.2em}

{\leftskip=1em
Apply Layer Normalization to the critic network.\par}
}

These two interventions are complementary and sufficient.
Data-anchored warmup ensures the critic begins with accurate value estimates in the relevant region, while normalization enables it to distinguish the subtle variations introduced by the residual policy.
Together, they establish the foundation for efficient value learning without requiring modifications to the underlying RL algorithm.

The power of \textbf{\texttt{DAWN}} lies not in novel components. 
Warmup data collection and normalization are both well-established techniques; our contribution lies in identifying \textit{why} and \textit{where} they become essential for residual value learning.
Complex alternatives such as carefully tuned distributional objectives, task-specific warmup data collection strategies, or sophisticated entropy scheduling during explicit warmup may yield marginal gains, but they do not address the core bottlenecks.
In contrast, the simplicity of \textbf{\texttt{DAWN}} is a direct consequence of targeting the fundamental challenges: once the critic is properly anchored and sensitive to residual variations, efficient value learning follows naturally. 
This simplicity also yields strong generality and robustness: \textbf{\texttt{DAWN}} introduces only two hyperparameters ($M$ and $\lambda$) and demonstrates consistent effectiveness across diverse tasks, base policies, and observation modalities, as shown in \Section~\ref{sec:experiments}.

\begin{figure*}[t]
\centering
\vspace{-0.6\baselineskip}
\begin{center}
\includegraphics[width=\linewidth]{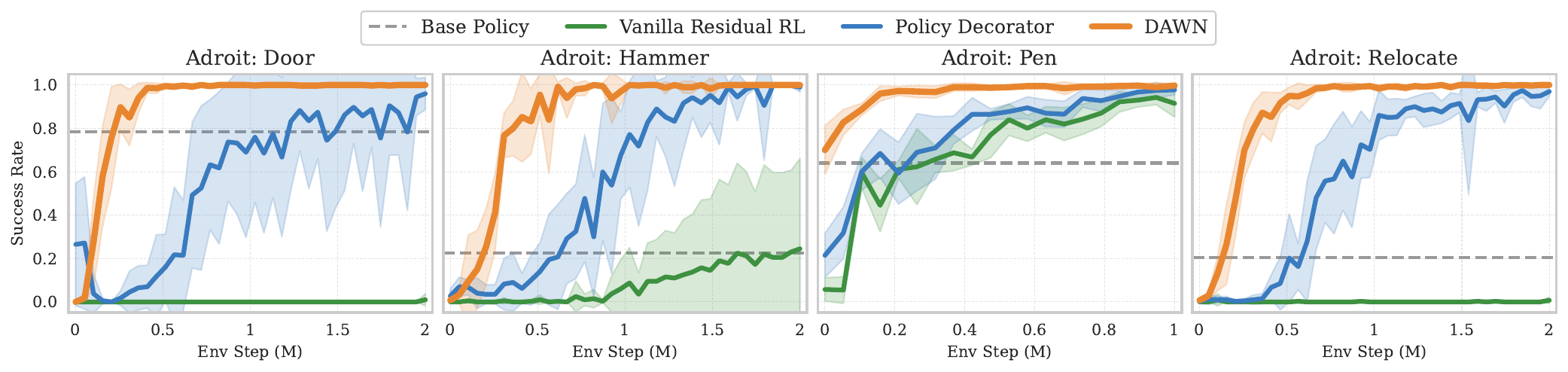}
\end{center}
\vspace{-0.8\baselineskip}
\caption{\textbf{Main results with BeT as the base policy.} \DAWN~maintains strong performance across all Adroit tasks, demonstrating robustness to different base policy architectures.
Shaded regions show 95\% confidence intervals over 8 seeds.}
\vspace{-\baselineskip}
\label{fig:bet}
\end{figure*}

\vspace{-0.1\baselineskip}
\section{Experiments}
\label{sec:experiments}

We assess whether insights from \Section~\ref{sec:dissecting} yield practical gains by evaluating \DAWN~across diverse settings.

\textbf{Setup.}~
We consider two benchmarks: ManiSkill~\citep{gu2023maniskill2} for contact-rich manipulation and Adroit~\citep{rajeswaran2017learning} for dexterous hand control.
To show that our insights apply beyond a specific base policy, we consider two distinct architectures: Diffusion Policy~\citep{chi2025diffusion} and Behavior Transformer (BeT)~\citep{shafiullah2022behavior}.
Full implementation details are provided in \Appendix~\ref{app:implementation}.

\textbf{Baselines.}~
Our primary baseline is \textit{Policy Decorator}~\citep{yuan2025policy}, the state-of-the-art residual RL method that introduces bounded residual actions and progressive exploration for stable learning. We also include \textit{Vanilla Residual RL}~\citep{johannink2019residual} as a minimal reference without stabilization mechanisms. 
Since Policy Decorator has been shown to substantially outperform methods such as JSRL~\citep{uchendu2023jump}, Cal-QL~\citep{nakamoto2023cal}, and FISH~\citep{haldar2023teach}, we focus our comparison on this stronger baseline.

\textbf{Main Results.}~
\Figure~\ref{fig:overall} presents results with Diffusion Policy as the base policy across 8 tasks spanning ManiSkill and Adroit benchmarks, including 2 tasks with visual observations; \Figure~\ref{fig:bet} further validates generalization to BeT as the base policy.
\DAWN~consistently achieves superior sample efficiency compared to all baselines, converging to high success rates significantly faster. 
On the challenging PegInsertionSide task, \DAWN~reaches 90\% success approximately $6\times$ faster than Policy Decorator, demonstrating that directly addressing the fundamental bottlenecks of value learning yields far greater efficiency gains than defensive mechanisms like progressive exploration.
Beyond efficiency, \DAWN~exhibits notably low variance across seeds on all tasks, suggesting that addressing the root causes of inefficient value learning also improves training stability. Such reliability is essential for real-world deployment.

\begin{figure}[t]
\centering
\vspace{-0.5\baselineskip}
\begin{center}
\includegraphics[width=\linewidth]{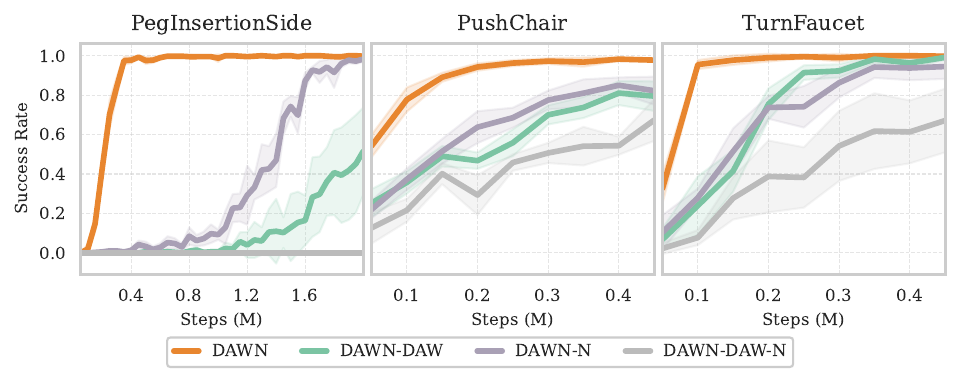}
\end{center}
\vspace{-\baselineskip}
\caption{\textbf{Ablation on \DAWN~components.} Data-anchored warmup (\textbf{\texttt{DAW}}) and critic normalization (\textbf{\texttt{N}}) address complementary bottlenecks. Removing either component degrades performance.}
\vspace{-1.5\baselineskip}
\label{fig:ablation_component}
\end{figure}

\begin{figure}[t]
\centering
\vspace{-0.6\baselineskip}
\begin{center}
\includegraphics[width=\linewidth]{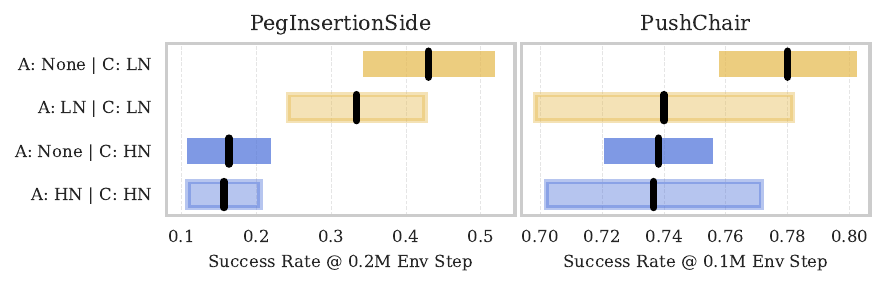}
\end{center}
\vspace{-0.9\baselineskip}
\caption{\textbf{Effect of actor \textit{vs.} critic normalization.} Normalization benefits the critic but not the actor, confirming that scale mismatch affects value learning rather than policy optimization.}
\vspace{-1.4\baselineskip}
\label{fig:ablation_actor}
\end{figure}

\textbf{Ablation.}~
\Figure~\ref{fig:ablation_component} confirms that both components of \DAWN~are essential: removing either data-anchored warmup or critic normalization substantially degrades learning efficiency.
As analyzed in \Section~\ref{sec:dissecting}, these two components target distinct bottlenecks and their benefits do not overlap.

We further examine whether normalization benefits the actor, the critic, or both. Adding normalization to the actor provides no improvement and may slightly degrade performance (\Figure~\ref{fig:ablation_actor}), consistent with our analysis that scale mismatch is primarily a value-side challenge.
Further ablations on hyperparameter sensitivity and design variations are provided in \Appendix~\ref{app:ablation}.
\vspace{-0.2\baselineskip}
\section{Conclusion}

This work investigated what makes value learning a bottleneck in residual RL and how to address it. By identifying \textit{cold start pathology} and \textit{scale mismatch} as the root causes, we developed \DAWN, a minimal approach that achieves substantial efficiency gains through two simple interventions: data-anchored warmup and critic normalization.

Beyond proposing a specific algorithm, this work provides mechanistic insight into value learning dynamics unique to residual RL.
For instance, we show that despite the residual being nearly invisible in action space, its effect on Q-values manifests as a clear mean shift, explaining both why standard critics struggle to assign credit and why capturing the full value distribution is unnecessary.
More broadly, this work illustrates a research philosophy: rather than introducing complex machinery to mask symptoms, tracing problems to their root causes enables solutions that are both simpler and more effective.

\textbf{Future Work.}~
While the identified bottlenecks and proposed solutions are general, extending this work to real-world robotic systems and larger vision-language-action models remains an important direction.
We hope this work encourages the community to consider the critical role of value learning in fine-tuning foundation policies.

\section*{Impact Statement}
This paper presents work whose goal is to advance the field
of Machine Learning. There are many potential societal
consequences of our work, none which we feel must be
specifically highlighted here.


\bibliography{ref}

@inproceedings{
lee2025hyperspherical,
title={Hyperspherical Normalization for Scalable Deep Reinforcement Learning},
author={Hojoon Lee and Youngdo Lee and Takuma Seno and Donghu Kim and Peter Stone and Jaegul Choo},
booktitle={Forty-second International Conference on Machine Learning},
year={2025},
url={https://openreview.net/forum?id=kfYxyvCYQ4}
}

@inproceedings{bellemare2017distributional,
  title={A distributional perspective on reinforcement learning},
  author={Bellemare, Marc G and Dabney, Will and Munos, R{\'e}mi},
  booktitle={International conference on machine learning},
  pages={449--458},
  year={2017},
  organization={PMLR}
}

@inproceedings{dabney2018distributional,
  title={Distributional reinforcement learning with quantile regression},
  author={Dabney, Will and Rowland, Mark and Bellemare, Marc and Munos, R{\'e}mi},
  booktitle={Proceedings of the AAAI conference on artificial intelligence},
  volume={32},
  number={1},
  year={2018}
}

@inproceedings{kuznetsov2020controlling,
  title={Controlling overestimation bias with truncated mixture of continuous distributional quantile critics},
  author={Kuznetsov, Arsenii and Shvechikov, Pavel and Grishin, Alexander and Vetrov, Dmitry},
  booktitle={International conference on machine learning},
  pages={5556--5566},
  year={2020},
  organization={PMLR}
}

@article{silver2018residual,
  title={Residual policy learning},
  author={Silver, Tom and Allen, Kelsey and Tenenbaum, Josh and Kaelbling, Leslie},
  journal={arXiv preprint arXiv:1812.06298},
  year={2018}
}

@inproceedings{johannink2019residual,
  title={Residual reinforcement learning for robot control},
  author={Johannink, Tobias and Bahl, Shikhar and Nair, Ashvin and Luo, Jianlan and Kumar, Avinash and Loskyll, Matthias and Ojea, Juan Aparicio and Solowjow, Eugen and Levine, Sergey},
  booktitle={2019 international conference on robotics and automation (ICRA)},
  pages={6023--6029},
  year={2019},
  organization={IEEE}
}

@article{black2024pi_0,
    title={\(\pi_0\): A Vision-Language-Action Flow Model for General Robot Control},
  author={Black, Kevin and Brown, Noah and Driess, Danny and Esmail, Adnan and Equi, Michael and Finn, Chelsea and Fusai, Niccolo and Groom, Lachy and Hausman, Karol and Ichter, Brian and others},
  journal={arXiv preprint arXiv:2410.24164},
  year={2024}
}

@article{kim2024openvla,
  title={Openvla: An open-source vision-language-action model},
  author={Kim, Moo Jin and Pertsch, Karl and Karamcheti, Siddharth and Xiao, Ted and Balakrishna, Ashwin and Nair, Suraj and Rafailov, Rafael and Foster, Ethan and Lam, Grace and Sanketi, Pannag and others},
  journal={arXiv preprint arXiv:2406.09246},
  year={2024}
}

@article{chi2025diffusion,
  title={Diffusion policy: Visuomotor policy learning via action diffusion},
  author={Chi, Cheng and Xu, Zhenjia and Feng, Siyuan and Cousineau, Eric and Du, Yilun and Burchfiel, Benjamin and Tedrake, Russ and Song, Shuran},
  journal={The International Journal of Robotics Research},
  volume={44},
  number={10-11},
  pages={1684--1704},
  year={2025},
  publisher={Sage Publications Sage UK: London, England}
}

@inproceedings{
yuan2025policy,
title={Policy Decorator: Model-Agnostic Online Refinement for Large Policy Model},
author={Xiu Yuan and Tongzhou Mu and Stone Tao and Yunhao Fang and Mengke Zhang and Hao Su},
booktitle={The Thirteenth International Conference on Learning Representations},
year={2025},
url={https://openreview.net/forum?id=e5jGTEiJMT}
}

@inproceedings{haarnoja2018soft,
  title={Soft actor-critic: Off-policy maximum entropy deep reinforcement learning with a stochastic actor},
  author={Haarnoja, Tuomas and Zhou, Aurick and Abbeel, Pieter and Levine, Sergey},
  booktitle={International conference on machine learning},
  pages={1861--1870},
  year={2018},
  organization={Pmlr}
}

@article{lillicrap2015continuous,
  title={Continuous control with deep reinforcement learning},
  author={Lillicrap, Timothy P and Hunt, Jonathan J and Pritzel, Alexander and Heess, Nicolas and Erez, Tom and Tassa, Yuval and Silver, David and Wierstra, Daan},
  journal={arXiv preprint arXiv:1509.02971},
  year={2015}
}

@article{ankile2025residual,
  title={Residual off-policy rl for finetuning behavior cloning policies},
  author={Ankile, Lars and Jiang, Zhenyu and Duan, Rocky and Shi, Guanya and Abbeel, Pieter and Nagabandi, Anusha},
  journal={arXiv preprint arXiv:2509.19301},
  year={2025}
}

@article{gu2023maniskill2,
  title={Maniskill2: A unified benchmark for generalizable manipulation skills},
  author={Gu, Jiayuan and Xiang, Fanbo and Li, Xuanlin and Ling, Zhan and Liu, Xiqiang and Mu, Tongzhou and Tang, Yihe and Tao, Stone and Wei, Xinyue and Yao, Yunchao and others},
  journal={arXiv preprint arXiv:2302.04659},
  year={2023}
}

@article{ba2016layer,
  title={Layer normalization},
  author={Ba, Jimmy Lei and Kiros, Jamie Ryan and Hinton, Geoffrey E},
  journal={arXiv preprint arXiv:1607.06450},
  year={2016}
}

@inproceedings{
lee2025simba,
title={SimBa: Simplicity Bias for Scaling Up Parameters in Deep Reinforcement Learning},
author={Hojoon Lee and Dongyoon Hwang and Donghu Kim and Hyunseung Kim and Jun Jet Tai and Kaushik Subramanian and Peter R. Wurman and Jaegul Choo and Peter Stone and Takuma Seno},
booktitle={The Thirteenth International Conference on Learning Representations},
year={2025},
url={https://openreview.net/forum?id=jXLiDKsuDo}
}

@article{seo2025fasttd3,
  title={FastTD3: Simple, Fast, and Capable Reinforcement Learning for Humanoid Control},
  author={Seo, Younggyo and Sferrazza, Carmelo and Geng, Haoran and Nauman, Michal and Yin, Zhao-Heng and Abbeel, Pieter},
  journal={arXiv preprint arXiv:2505.22642},
  year={2025}
}

@inproceedings{
nauman2025bigger,
title={Bigger, Regularized, Categorical: High-Capacity Value Functions are Efficient Multi-Task Learners},
author={Michal Nauman and Marek Cygan and Carmelo Sferrazza and Aviral Kumar and Pieter Abbeel},
booktitle={The Thirty-ninth Annual Conference on Neural Information Processing Systems},
year={2025},
url={https://openreview.net/forum?id=zhOUfuOIzA}
}

@misc{
anonymous2026unified,
title={Unified Latent Steering and Residual Refinement for Online Improvement of Diffusion Policy Models},
author={Anonymous},
year={2026},
url={https://openreview.net/forum?id=DbBD2aT1OG}
}

@article{nakamoto2023cal,
  title={Cal-ql: Calibrated offline rl pre-training for efficient online fine-tuning},
  author={Nakamoto, Mitsuhiko and Zhai, Simon and Singh, Anikait and Sobol Mark, Max and Ma, Yi and Finn, Chelsea and Kumar, Aviral and Levine, Sergey},
  journal={Advances in Neural Information Processing Systems},
  volume={36},
  pages={62244--62269},
  year={2023}
}

@article{intelligence2025pi,
  title={$\pi^{*}_{0.6}$: a VLA That Learns From Experience},
  author={Intelligence, Physical and Amin, Ali and Aniceto, Raichelle and Balakrishna, Ashwin and Black, Kevin and Conley, Ken and Connors, Grace and Darpinian, James and Dhabalia, Karan and DiCarlo, Jared and others},
  journal={arXiv preprint arXiv:2511.14759},
  year={2025}
}

@article{song2025survey,
  title={A survey on diffusion policy for robotic manipulation: Taxonomy, analysis, and future directions},
  author={Song, Mingchen and Deng, Xiang and Zhou, Zhiling and Wei, Jie and Guan, Weili and Nie, Liqiang},
  journal={Authorea Preprints},
  year={2025},
  publisher={Authorea}
}

@article{lei2025rl,
  title={Rl-100: Performant robotic manipulation with real-world reinforcement learning},
  author={Lei, Kun and Li, Huanyu and Yu, Dongjie and Wei, Zhenyu and Guo, Lingxiao and Jiang, Zhennan and Wang, Ziyu and Liang, Shiyu and Xu, Huazhe},
  journal={arXiv preprint arXiv:2510.14830},
  year={2025}
}

@article{li2025vla,
  title={Vla-rft: Vision-language-action reinforcement fine-tuning with verified rewards in world simulators},
  author={Li, Hengtao and Ding, Pengxiang and Suo, Runze and Wang, Yihao and Ge, Zirui and Zang, Dongyuan and Yu, Kexian and Sun, Mingyang and Zhang, Hongyin and Wang, Donglin and others},
  journal={arXiv preprint arXiv:2510.00406},
  year={2025}
}

@article{silver2025welcome,
  title={Welcome to the era of experience},
  author={Silver, David and Sutton, Richard S},
  journal={Google AI},
  volume={1},
  year={2025}
}

@article{pan2026sop,
  title={SOP: A Scalable Online Post-Training System for Vision-Language-Action Models},
  author={Pan, Mingjie and Feng, Siyuan and Zhang, Qinglin and Li, Xinchen and Song, Jianheng and Qu, Chendi and Wang, Yi and Li, Chuankang and Xiong, Ziyu and Chen, Zhi and others},
  journal={arXiv preprint arXiv:2601.03044},
  year={2026}
}

@inproceedings{ankile2025imitation,
  title={From imitation to refinement-residual rl for precise assembly},
  author={Ankile, Lars and Simeonov, Anthony and Shenfeld, Idan and Torne, Marcel and Agrawal, Pulkit},
  booktitle={2025 IEEE International Conference on Robotics and Automation (ICRA)},
  pages={01--08},
  year={2025},
  organization={IEEE}
}

@article{xiao2025self,
  title={Self-improving vision-language-action models with data generation via residual rl},
  author={Xiao, Wenli and Lin, Haotian and Peng, Andy and Xue, Haoru and He, Tairan and Xie, Yuqi and Hu, Fengyuan and Wu, Jimmy and Luo, Zhengyi and Fan, Linxi and others},
  journal={arXiv preprint arXiv:2511.00091},
  year={2025}
}

@inproceedings{
jiang2024transic,
title={{TRANSIC}: Sim-to-Real Policy Transfer by Learning from Online Correction},
author={Yunfan Jiang and Chen Wang and Ruohan Zhang and Jiajun Wu and Li Fei-Fei},
booktitle={8th Annual Conference on Robot Learning},
year={2024},
url={https://openreview.net/forum?id=lpjPft4RQT}
}

@inproceedings{
zhou2025efficient,
title={Efficient Online Reinforcement Learning Fine-Tuning Need Not Retain Offline Data},
author={Zhiyuan Zhou and Andy Peng and Qiyang Li and Sergey Levine and Aviral Kumar},
booktitle={The Thirteenth International Conference on Learning Representations},
year={2025},
url={https://openreview.net/forum?id=HN0CYZbAPw}
}

@inproceedings{
wang2025residualmppi,
title={Residual-{MPPI}: Online Policy Customization for Continuous Control},
author={Pengcheng Wang and Chenran Li and Catherine Weaver and Kenta Kawamoto and Masayoshi Tomizuka and Chen Tang and Wei Zhan},
booktitle={The Thirteenth International Conference on Learning Representations},
year={2025},
url={https://openreview.net/forum?id=gVnJFY8nCM}
}

@article{rajeswaran2017learning,
  title={Learning complex dexterous manipulation with deep reinforcement learning and demonstrations},
  author={Rajeswaran, Aravind and Kumar, Vikash and Gupta, Abhishek and Vezzani, Giulia and Schulman, John and Todorov, Emanuel and Levine, Sergey},
  journal={arXiv preprint arXiv:1709.10087},
  year={2017}
}

@article{shafiullah2022behavior,
  title={Behavior transformers: Cloning $ k $ modes with one stone},
  author={Shafiullah, Nur Muhammad and Cui, Zichen and Altanzaya, Ariuntuya Arty and Pinto, Lerrel},
  journal={Advances in neural information processing systems},
  volume={35},
  pages={22955--22968},
  year={2022}
}

@article{mu2021maniskill,
  title={Maniskill: Generalizable manipulation skill benchmark with large-scale demonstrations},
  author={Mu, Tongzhou and Ling, Zhan and Xiang, Fanbo and Yang, Derek and Li, Xuanlin and Tao, Stone and Huang, Zhiao and Jia, Zhiwei and Su, Hao},
  journal={arXiv preprint arXiv:2107.14483},
  year={2021}
}

@inproceedings{uchendu2023jump,
  title={Jump-start reinforcement learning},
  author={Uchendu, Ikechukwu and Xiao, Ted and Lu, Yao and Zhu, Banghua and Yan, Mengyuan and Simon, Jos{\'e}phine and Bennice, Matthew and Fu, Chuyuan and Ma, Cong and Jiao, Jiantao and others},
  booktitle={International Conference on Machine Learning},
  pages={34556--34583},
  year={2023},
  organization={PMLR}
}

@article{haldar2023teach,
  title={Teach a robot to fish: Versatile imitation from one minute of demonstrations},
  author={Haldar, Siddhant and Pari, Jyothish and Rai, Anant and Pinto, Lerrel},
  journal={arXiv preprint arXiv:2303.01497},
  year={2023}
}

@inproceedings{ball2023efficient,
  title={Efficient online reinforcement learning with offline data},
  author={Ball, Philip J and Smith, Laura and Kostrikov, Ilya and Levine, Sergey},
  booktitle={International Conference on Machine Learning},
  pages={1577--1594},
  year={2023},
  organization={PMLR}
}

@InProceedings{Xiang_2020_SAPIEN,
author = {Xiang, Fanbo and Qin, Yuzhe and Mo, Kaichun and Xia, Yikuan and Zhu, Hao and Liu, Fangchen and Liu, Minghua and Jiang, Hanxiao and Yuan, Yifu and Wang, He and Yi, Li and Chang, Angel X. and Guibas, Leonidas J. and Su, Hao},
title = {{SAPIEN}: A SimulAted Part-based Interactive ENvironment},
booktitle = {The IEEE Conference on Computer Vision and Pattern Recognition (CVPR)},
month = {June},
year = {2020}}

@inproceedings{zitkovich2023rt,
  title={Rt-2: Vision-language-action models transfer web knowledge to robotic control},
  author={Zitkovich, Brianna and Yu, Tianhe and Xu, Sichun and Xu, Peng and Xiao, Ted and Xia, Fei and Wu, Jialin and Wohlhart, Paul and Welker, Stefan and Wahid, Ayzaan and others},
  booktitle={Conference on Robot Learning},
  pages={2165--2183},
  year={2023},
  organization={PMLR}
}

@article{xu2025compliant,
  title={Compliant Residual DAgger: Improving Real-World Contact-Rich Manipulation with Human Corrections},
  author={Xu, Xiaomeng and Hou, Yifan and Liu, Zeyi and Song, Shuran},
  journal={arXiv preprint arXiv:2506.16685},
  year={2025}
}

@article{yan2026efficiently,
  title={Efficiently Learning Robust Torque-based Locomotion Through Reinforcement with Model-Based Supervision},
  author={Yan, Yashuai and Egle, Tobias and Ott, Christian and Lee, Dongheui},
  journal={arXiv preprint arXiv:2601.16109},
  year={2026}
}

@article{ren2024dppo,
  title={Diffusion policy policy optimization},
  author={Ren, Allen Z and Lidard, Justin and Ankile, Lars L and Simeonov, Anthony and Agrawal, Pulkit and Majumdar, Anirudha and Burchfiel, Benjamin and Dai, Hongkai and Simchowitz, Max},
  journal={arXiv preprint arXiv:2409.00588},
  year={2024}
}

@article{zhang2025reinflow,
  title={ReinFlow: Fine-tuning flow matching policy with online reinforcement learning},
  author={Zhang, Tonghe and Yu, Chao and Su, Sichang and Wang, Yu},
  journal={arXiv preprint arXiv:2505.22094},
  year={2025}
}

@article{mark2024parl,
  title={Policy agnostic rl: Offline rl and online rl fine-tuning of any class and backbone},
  author={Mark, Max Sobol and Gao, Tian and Sampaio, Georgia Gabriela and Srirama, Mohan Kumar and Sharma, Archit and Finn, Chelsea and Kumar, Aviral},
  journal={arXiv preprint arXiv:2412.06685},
  year={2024}
}

@article{wagenmaker2025dsrl,
  title={Steering Your Diffusion Policy with Latent Space Reinforcement Learning},
  author={Wagenmaker, Andrew and Nakamoto, Mitsuhiko and Zhang, Yunchu and Park, Seohong and Yagoub, Waleed and Nagabandi, Anusha and Gupta, Abhishek and Levine, Sergey},
  journal={arXiv preprint arXiv:2506.15799},
  year={2025}
}
\bibliographystyle{icml2026}

\newpage
\appendix
\onecolumn

\definecolor{lightgraybg}{gray}{0.95}

\newcommand{\partseparator}{%
    \par            
    \addvspace{0.2em} 
    \noindent       
    {\color{gray!40}\rule{\linewidth}{0.5pt}}
    \par            
    \vspace{0.3em}  
    \noindent       
}

\vspace*{0.1ex}
\begin{tcolorbox}[
    colback=lightgraybg,
    colframe=lightgraybg,
    boxrule=0pt,
    arc=6pt,
    boxsep=8pt,
    left=0pt, right=0pt,
    top=1pt, bottom=3pt,
    halign=center,
    width=\linewidth,
    after skip=0.4in
]
    {\large \bfseries Appendix for} \\[0.8em]
    {\large \itshape What Makes Value Learning Efficient in Residual Reinforcement Learning?}
\end{tcolorbox}


\partseparator
{\large \bfseries \textcolor{black}{Part I: Background and Setup} \par} 
\startcontents[part1]
\printcontents[part1]{}{1}{\setcounter{tocdepth}{2}}
\stopcontents[part1]

\partseparator
{\large \bfseries \textcolor{black}{Part II: Investigation Details} \par}
\startcontents[part2]
\printcontents[part2]{}{1}{\setcounter{tocdepth}{2}}
\stopcontents[part2]

\partseparator
{\large \bfseries \textcolor{black}{Part III: Method and Ablations} \par}
\startcontents[part3]
\printcontents[part3]{}{1}{\setcounter{tocdepth}{2}}
\stopcontents[part3]

\partseparator


\resumecontents[part1]
\newpage
\section{Related Work}
\label{app:related}

\subsection{Residual Reinforcement Learning}
\label{app:related_residual}

Residual RL learns a correction term on top of a fixed base policy, preserving prior knowledge while adapting to unmodeled dynamics.
Originally applied to classical controllers in contact-rich robotics tasks~\citep{silver2018residual, johannink2019residual, haldar2023teach, yan2026efficiently}, the paradigm has recently gained traction for online refinement of large pretrained policies, including behavior cloning~\citep{ankile2025imitation}, diffusion policies~\citep{anonymous2026unified}, and vision-language-action models~\citep{xiao2025self}, where direct fine-tuning is sample-inefficient and risks catastrophic forgetting.

Prior work has addressed training challenges primarily through policy-side mechanisms: action bounding, progressive exploration schedules, and controlled exploration strategies~\citep{silver2018residual, yuan2025policy, ankile2025residual, xu2025compliant}.
These techniques treat symptoms of instability rather than root causes.
In contrast, we study residual RL from a \textit{value learning} perspective, identifying critic cold start and scale mismatch as the fundamental bottlenecks, and develop principled fixes that address these causes directly.

\textit{\textbf{Concurrent Work.}}~
ResFiT~\citep{ankile2025residual} is an impressive concurrent work that demonstrates residual RL on real-world high-DoF humanoid systems.
Their ablations show the effectiveness of critic normalization following RLPD~\citep{ball2023efficient}; our work complements this by providing mechanistic analysis of \textit{why} normalization restores critic sensitivity in the residual setting.
Additionally, while ResFiT leverages offline demonstrations with reward labels, we show that stable residual RL can be achieved with only base policy rollouts serving as a value anchor, which is a more minimal requirement that may broaden practical applicability.

\subsection{Fine-tuning Foundation Policies via RL}
\label{app:related_finetuning}

Large pretrained policies such as diffusion policies~\citep{chi2025diffusion}, behavior transformers~\citep{shafiullah2022behavior}, and vision-language-action models~\citep{zitkovich2023rt} have become powerful foundations for robot learning but often underperform in deployment due to distribution shift.
RL fine-tuning offers a natural remedy, yet existing approaches face distinct challenges depending on whether they modify the base policy.

Direct fine-tuning methods~\citep{ren2024dppo, zhang2025reinflow, mark2024parl} update the base policy parameters via policy gradients.
While effective, these approaches must carefully manage training stability, catastrophic forgetting, and architecture-specific optimization challenges.
In contrast, residual methods~\citep{yuan2025policy, ankile2025imitation, ankile2025residual, wagenmaker2025dsrl} freeze the base policy and learn only a lightweight correction, offering two key advantages: \textit{stability} by preserving the base policy's knowledge, and \textit{model-agnosticism} by applying regardless of base architecture.
A separate line of work from the offline-to-online RL literature~\citep{nakamoto2023cal, ball2023efficient, uchendu2023jump} accelerates fine-tuning by leveraging offline data to initialize critics or guide exploration.
However, these methods typically assume access to offline datasets with reward annotations or require specific algorithmic machinery (e.g., conservative Q-learning), limiting their applicability when only a pretrained policy is available.

Our work uses residual RL as a controlled setting to study value learning dynamics.
We show that targeted improvements to the critic can substantially boost efficiency, suggesting that a deeper understanding of RL optimization fundamentals may complement ongoing efforts in policy fine-tuning.
\newpage
\section{Experimental Setup}

\subsection{Environment Details}
\label{app:environment}

We evaluate on several tasks from two complementary benchmarks that together cover a broad spectrum of manipulation challenges: ManiSkill~\citep{mu2021maniskill, gu2023maniskill2} for contact-rich manipulation, and Adroit~\citep{rajeswaran2017learning} for high-dimensional dexterous control with an anthropomorphic hand.

\paragraph{ManiSkill.}
ManiSkill is a large-scale benchmark for generalizable manipulation skills built on the SAPIEN simulator~\citep{Xiang_2020_SAPIEN}.
It provides diverse task families with object-level geometric and topological variations, making it well-suited for studying generalization in manipulation.
We select three challenging tasks that stress different aspects of residual learning:
\begin{itemize}[leftmargin=*,itemsep=2pt,topsep=2pt]
    \item \textbf{PegInsertionSide}: Insert a peg into a horizontal hole with randomized box geometry. This task demands sub-millimeter precision and is highly sensitive to policy errors—making it an ideal testbed for measuring the efficiency of value learning.
    \item \textbf{TurnFaucet}: Rotate a faucet handle to turn it on. The base policy is trained on 10 faucet instances, while online evaluation uses 4 \textit{unseen} faucets, testing generalization to novel object geometries.
    \item \textbf{PushChair}: A dual-arm mobile manipulator must push a swivel chair to a target location without tipping it over. Trained on 5 chair instances and evaluated on 3 unseen chairs with randomized friction and damping parameters.
\end{itemize}
All ManiSkill tasks use a Franka Panda robot (or dual-arm variant for PushChair) with delta end-effector control.
For visual observation experiments, we use 64 $\times$ 64 RGBD images from base and hand-mounted cameras.

\paragraph{Adroit.}
The Adroit benchmark~\citep{rajeswaran2017learning} features the Shadow Dexterous Hand, a 24-DoF anthropomorphic hand mounted on a 4--6 DoF arm, totaling 28--30 DoF depending on the task.
Originally introduced to study demonstration-augmented policy gradient (DAPG), these tasks remain challenging benchmarks for dexterous manipulation due to their high-dimensional action spaces and contact-rich dynamics.
We evaluate on four tasks:
\begin{itemize}[leftmargin=*,itemsep=2pt,topsep=2pt]
    \item \textbf{Door}: Undo a latch with significant dry friction and bias torque, then swing the door open. The door position is randomized at each episode.
    \item \textbf{Pen}: Reorient a pen to match a randomized target configuration, requiring precise in-hand manipulation.
    \item \textbf{Hammer}: Grasp a hammer and drive a nail into a board. The nail position is randomized and exerts up to 15N of resistive friction.
    \item \textbf{Relocate}: Move a ball to a target location, with both positions randomized across the workspace.
\end{itemize}
All Adroit tasks use absolute joint position control.

\paragraph{Task Specifications.}
\Table~\ref{tab:task_specs} summarizes the observation and action dimensions for each task.
The high action dimensionality of Adroit tasks (24--30 DoF) presents a distinct challenge compared to the lower-dimensional but precision-critical ManiSkill tasks.

\begin{table}[ht]
\centering
\caption{Task specifications across benchmarks. ManiSkill tasks feature lower action dimensions but require high precision; Adroit tasks involve high-dimensional dexterous control.}
\label{tab:task_specs}
\begin{tabular}{lcccc}
\toprule
Task & State Dim & Action Dim & Episode Length \\
\midrule
ManiSkill: PegInsertionSide & 50 & 7 & 200 \\
ManiSkill: TurnFaucet & 43 & 7 & 200 \\
ManiSkill: PushChair & 131 & 20 & 200 \\
\midrule
Adroit: Door & 39 & 28 & 300 \\
Adroit: Pen & 46 & 24 & 200 \\
Adroit: Hammer & 46 & 26 & 400 \\
Adroit: Relocate & 39 & 30 & 400 \\
\bottomrule
\end{tabular}
\end{table}

\subsection{Base Policy Training}
\label{app:base_policy}

\textit{\textbf{To ensure fair comparison with prior work, we use the pretrained base policy checkpoints released by \citet{yuan2025policy} rather than training our own.}}
This section documents the base policy architectures and training procedures for reference; readers primarily interested in our method can skip to Section~\ref{app:baseline}.

We experiment with two expressive policy architectures: Diffusion Policy~\citep{chi2025diffusion} and Behavior Transformer (BeT)~\citep{shafiullah2022behavior}.
Diffusion Policy models the action distribution as a conditional denoising process and has demonstrated strong performance on contact-rich manipulation tasks.
BeT uses a transformer architecture with action discretization via k-means clustering, enabling multimodal action prediction.

\paragraph{Diffusion Policy.}
We use the U-Net variant of Diffusion Policy.
Key architecture hyperparameters are listed in Table~\ref{tab:diffusion_arch}.
The policy observes 2 timesteps of history and predicts 16 future actions, of which 4 are executed before re-planning.

\begin{table}[h]
\centering
\caption{Diffusion Policy architecture.}
\label{tab:diffusion_arch}
\begin{tabular}{lc}
\toprule
Hyperparameter & Value \\
\midrule
Observation horizon & 2 \\
Prediction horizon & 16 \\
Action horizon & 4 \\
Embedding dimension & 64 \\
Down-sampling dimensions & 256, 512, 1024 \\
Parameters & $\sim$4M \\
\bottomrule
\end{tabular}
\end{table}

\paragraph{Behavior Transformer.}
BeT discretizes the action space using k-means clustering and models action sequences autoregressively.
Architecture hyperparameters are listed in Table~\ref{tab:bet_arch}.

\begin{table}[h]
\centering
\caption{Behavior Transformer architecture.}
\label{tab:bet_arch}
\begin{tabular}{lc}
\toprule
Hyperparameter & Value \\
\midrule
Context window & 10 / 20 \\
Number of clusters & 4 / 8 \\
Transformer layers & 4 \\
Attention heads & 4 \\
Embedding dimension & 128 \\
Parameters & $\sim$1M \\
\bottomrule
\end{tabular}
\end{table}

\paragraph{Training.}
Both policies are trained using the AdamW optimizer with a learning rate of $10^{-4}$.
Diffusion Policy is trained for 200K gradient steps with batch size 1024 on all tasks.
BeT is trained for 200K steps on ManiSkill tasks and 5K steps on Adroit tasks (which have fewer demonstrations), both with batch size 2048.
Checkpoints are selected based on the highest evaluation success rate over 50 episodes, evaluated every 50K steps.

\paragraph{Demonstration Data.}
Table~\ref{tab:demos} summarizes the demonstration data used for base policy training.
ManiSkill demonstrations are generated via motion planning or model predictive control and contain 1000 trajectories per task.
Adroit demonstrations are collected via human teleoperation with only 25 trajectories per task, reflecting the original DAPG setup~\citep{rajeswaran2017learning}.

\begin{table}[h]
\centering
\caption{Demonstration data for base policy training.}
\label{tab:demos}
\begin{tabular}{lcc}
\toprule
Benchmark & Trajectories & Collection Method \\
\midrule
ManiSkill & 1000 & Motion planning / MPC \\
Adroit & 25 & Human teleoperation \\
\bottomrule
\end{tabular}
\end{table}

\subsection{Details of Minimal Baseline}
\label{app:baseline}

Our implementation builds on \textbf{Policy Decorator}~\citep{yuan2025policy}, which introduces two strategies for stable residual learning. 
The first is \textit{bounded residual action}: scaling the residual output by a small $\lambda$ (as in Equation~\ref{eq:residual_action}) to prevent large deviations from the base policy. 
The second is \textit{progressive exploration}: gradually increasing the probability of using the residual policy during training,
\begin{equation}
\pi_{\text{behavior}}(s) = 
\begin{cases}
\pi_{\text{base}}(s) + \lambda \cdot \pi_{\text{res}}(s) & \text{if } u < \epsilon \\
\pi_{\text{base}}(s) & \text{otherwise}
\end{cases}
\label{eq:progressive}
\end{equation}
where $u \sim \text{Uniform}(0,1)$ and $\epsilon = \min(t/H, 1)$ increases linearly until timestep $H$, a task-specific hyperparameter that varies widely across tasks.

We adopt bounded residual action with the task-specific $\lambda$ values from~\citet{yuan2025policy}. 
However, we deliberately omit progressive exploration for two reasons:
\begin{enumerate}[leftmargin=*,itemsep=2pt,topsep=2pt]
    \item As a protective mechanism, progressive exploration masks the symptoms of inefficient value learning, obscuring the dynamics we aim to study.
    \item Once value learning is efficient, such protection becomes unnecessary and potentially harmful: the changing $\epsilon$ introduces training non-stationarity, and the mixture of base-only and full-policy rollouts creates distribution mismatch between data collection and value estimation.
\end{enumerate}
This minimal setup exposes the raw challenges of value learning and serves as our experimental subject in Section~\ref{sec:dissecting}.

\paragraph{Off-Policy RL Algorithm.}
We use Soft Actor-Critic (SAC)~\citep{haarnoja2018soft} as the underlying off-policy algorithm, following the standard practice in residual RL~\citep{yuan2025policy, ankile2025residual}.
SAC maintains two value networks $Q_{\theta_1}, Q_{\theta_2}$ along with EMA-updated target networks $Q_{\theta'_1}, Q_{\theta'_2}$, optimized by minimizing $\mathbb{E}[(Q_{\theta_i}(s,a) - y)^2]$ with the soft Bellman target:
\begin{equation}
y = r + \gamma \left( \min_{i \in \{1,2\}} Q_{\theta'_i}(s', a') - \alpha \log \pi(a' \mid s') \right),
\label{eq:soft_bellman}
\end{equation}
where $\gamma$ is the discount factor and $\alpha$ is the temperature for entropy regularization.
The entropy term encourages exploration, but interacts with the residual policy initialization in ways we analyze in Section~\ref{sec:explicit_warmup}: the near-deterministic initialization makes $|\log \pi|$ extremely large, causing pathological dynamics during explicit warmup.

\paragraph{Network Architecture.}
The residual policy $\pi_{\text{res}}$ is a lightweight MLP with 2 hidden layers of 256 units each, using ReLU activations.
The critic $Q_\theta$ follows the same architecture but takes the concatenation of state and \textit{summed action} $a = a_{\text{base}} + \lambda \cdot a_{\text{res}}$ as input.
We use twin critics to mitigate overestimation bias.
To avoid disrupting the base policy at initialization, the \textbf{residual policy's final layer is initialized with near-zero weights}, producing near-zero actions with minimal variance at the start of training.

\paragraph{Hyperparameters.}
Table~\ref{tab:sac_hp} lists the SAC hyperparameters shared across all experiments.
Table~\ref{tab:lambda_investigation} lists the residual scale $\lambda$ for the ManiSkill tasks used in Section~\ref{sec:dissecting}; full $\lambda$ values for all tasks are provided in Section~\ref{app:implementation}.

\begin{table}[h]
\centering
\begin{minipage}[t]{0.54\textwidth}
    \centering
    \caption{SAC hyperparameters.}
    \label{tab:sac_hp}
    \begin{tabular}{lcc}
    \toprule
    Hyperparameter & ManiSkill & Adroit \\
    \midrule
    Discount $\gamma$ & 0.97 / 0.9 & 0.97 \\
    Batch size & 1024 & 1024 \\
    Learning rate & $10^{-4}$ & $10^{-4}$ \\
    Initial entropy $\alpha$ & 0.01 & 0.01 \\
    Policy update frequency & 1 & 1 \\
    Env steps per update & 64 & 64 \\
    Update-to-data ratio & 0.25 & 0.25 \\
    Target network $\tau$ & 0.01 & 0.01 \\
    Replay buffer size & $10^6$ & $10^6$ \\
    \bottomrule
    \end{tabular}
\end{minipage}
\hfill
\begin{minipage}[t]{0.45\textwidth}
    \centering
    \caption{Residual scale $\lambda$ for investigation tasks (Section~\ref{sec:dissecting}).}
    \label{tab:lambda_investigation}
    \begin{tabular}{lc}
    \toprule
    Task & $\lambda$ \\
    \midrule
    PegInsertionSide & 0.1 \\
    TurnFaucet & 0.1 \\
    PushChair & 0.2 \\
    \bottomrule
    \end{tabular}
\end{minipage}
\end{table}

Note that standard off-policy RL implementations include a ``learning starts'' parameter that specifies the number of transitions to collect before training begins.
The original Policy Decorator uses 8K transitions for this purpose.
Our investigation in Section~\ref{sec:warmup_anchor} reveals that this warmup data serves as a critical value anchor for residual RL, and we systematically study the effect of varying this quantity from 0K to 40K.
Based on these findings, we adopt 20K warmup transitions as the default for \DAWN.
\stopcontents[part1]

\resumecontents[part2]
\section{Investigation of Cold Start Pathology}

\subsection{Implicit Warmup}
\label{app:implicit_warmup}

This section provides additional details and analysis on the implicit warmup mechanism discussed in Section~\ref{sec:warmup_anchor}.

\subsubsection{Warmup Data Collection Strategy}
\label{app:warmup_strategy}

Section~\ref{sec:warmup_anchor} establishes that warmup data is necessary for efficient value learning. 
A natural follow-up question arises: \textit{how should this warmup data be collected?}
One might expect that adding exploration during warmup would improve state-action coverage and thereby accelerate subsequent learning.
To test this hypothesis, we compare four collection strategies:

\begin{itemize}[leftmargin=*,itemsep=2pt,topsep=2pt]
    \item \textbf{Base Policy Only}: Collect transitions using $\pi_{\text{base}}$ alone, without any exploration mechanism. This is the simplest strategy and the one we adopt by default.
    
    \item \textbf{Full Action}: Execute the full combined policy $\pi_{\text{base}} + \lambda \cdot \pi_{\text{res}}$ with the randomly initialized residual. Since $\pi_{\text{res}}$ outputs near-zero actions at initialization, this introduces only minimal perturbations to the base policy's behavior.
    
    \item \textbf{Gaussian Noise}: Add isotropic Gaussian noise to the base policy actions, i.e., $a = \pi_{\text{base}}(s) + \epsilon$ where $\epsilon \sim \mathcal{N}(0, \sigma^2 I)$. We set $\sigma = 0.1$ to match the typical residual scale $\lambda$.
    
    \item \textbf{Epsilon-Greedy}: Mix base-only and full-policy rollouts following Equation~\ref{eq:progressive} with a fixed ratio $\epsilon = 0.2$, where 20\% of transitions use the combined policy and 80\% use the base policy alone.
\end{itemize}

\begin{figure}[ht]
\centering
\includegraphics[width=0.75\linewidth]{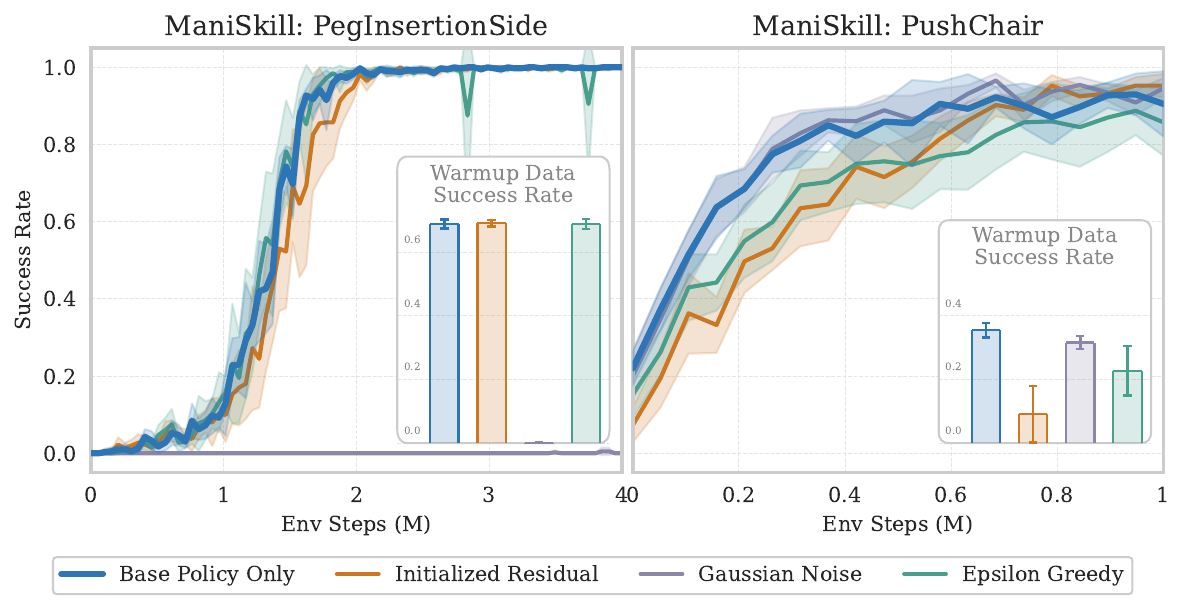}
\caption{\textbf{Comparison of warmup data collection strategies.} 
\textit{Base Policy Only} achieves comparable or better performance than strategies with additional exploration.
The inset bar charts show success rates during warmup data collection; exploration degrades warmup performance in a task-dependent manner.}
\label{fig:exploration}
\end{figure}

\paragraph{Results.}
Figure~\ref{fig:exploration} compares these strategies across two ManiSkill tasks.
Contrary to the intuition that more exploration should help, \textit{Base Policy Only} performs as well as or better than all exploration-augmented strategies.
The failure modes are task-dependent: Gaussian Noise causes PegInsertionSide to collapse entirely, while PushChair exhibits higher variance across strategies but remains more robust to perturbations.

\paragraph{Analysis.}
This result reinforces the \textit{value anchor} interpretation of warmup data.
The purpose of warmup is not to pre-collect diverse trajectories that cover a wide state-action space, but rather to ground the critic to accurate value estimates in the region where the base policy operates.
Since residual RL refines the base policy through local corrections rather than global policy search, the critic must first understand the value landscape around $\pi_{\text{base}}$ before it can meaningfully guide residual learning.
Adding exploration during warmup dilutes this grounding signal with transitions from regions that may be irrelevant or misleading for the initial phase of learning.

Once RL training begins, exploration emerges naturally through two mechanisms: (1) the stochastic residual policy $\pi_{\text{res}}$, which samples actions from a learned Gaussian distribution, and (2) the entropy regularization in SAC, which explicitly encourages exploration.
TD learning then propagates value estimates outward from the anchored foundation established by warmup data.
Our results do not rule out the possibility that more sophisticated exploration strategies could be beneficial; designing active exploration methods that specifically target value-critical regions for anchoring remains a promising direction for future work.
Here, we demonstrate that the simplest approach—collecting data from the base policy alone—already provides substantial benefits and offers a clear mechanistic explanation.

\paragraph{Trajectory-Consistent Noise: A Negative Result.}
The failure of step-wise Gaussian noise raises a natural question: is the problem with noise itself, or with its high-frequency nature?
In contact-rich manipulation tasks, the dynamics exhibit significant inertia and friction.
Step-wise white noise is largely filtered out by the physics, producing trajectories that stay close to the base behavior despite the added perturbations.
When the noise magnitude is increased to overcome this filtering, the robot quickly diverges from reasonable behavior and the success rate drops to near zero.

To test whether \textit{trajectory-consistent} noise could address this issue, we designed a more structured exploration strategy.
Instead of sampling independent noise at each timestep, we sample a single bias vector $b \sim \mathcal{N}(0, \sigma^2 I)$ at the start of each episode and apply it consistently throughout:
\begin{equation}
a^{\text{res}}_t = b + \epsilon_t, \quad \text{where } \epsilon_t \sim \mathcal{N}(0, \sigma_{\text{jitter}}^2 I)
\end{equation}
with $\sigma = 0.2$ for the trajectory-level bias and $\sigma_{\text{jitter}} = 0.01$ for small per-step jitter.
This design ensures that the effect of the residual integrates over time, producing clear differences in observed returns rather than being filtered out by the dynamics.

We compare two variants:
\begin{itemize}[leftmargin=*,itemsep=2pt,topsep=2pt]
    \item \textbf{TC Noise}: All trajectories use trajectory-consistent noise as described above.
    \item \textbf{TC Noise w/ Anchor}: Half of the trajectories use the pure base policy (anchor), and half use trajectory-consistent noise.
\end{itemize}

\begin{figure}[ht]
\centering
\includegraphics[width=\linewidth]{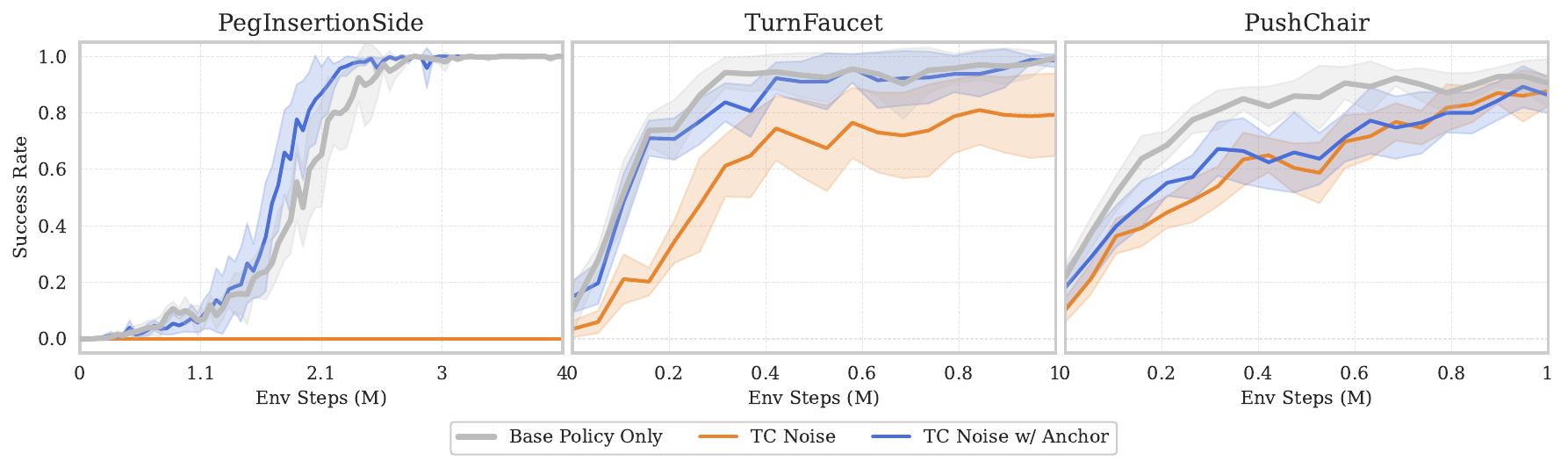}
\caption{\textbf{Trajectory-consistent noise does not improve over base policy only.}
TC Noise fails entirely on PegInsertionSide and underperforms on TurnFaucet.
Adding anchor trajectories (TC Noise w/ Anchor) recovers performance but provides no benefit over the simpler Base Policy Only strategy.}
\label{fig:tc_noise}
\end{figure}

Despite the more principled design, neither variant outperforms the simple \textit{Base Policy Only} strategy:
\begin{itemize}[leftmargin=*,itemsep=2pt,topsep=2pt]
    \item \textbf{TC Noise} performs substantially worse than \textit{Base Policy Only}, confirming that exploration during warmup—even when carefully structured—dilutes the value anchor signal.
    \item \textbf{TC Noise w/ Anchor} performs comparably to or slightly worse than \textit{Base Policy Only}. Adding anchor trajectories recovers most of the performance, but the probe trajectories provide no additional benefit.
\end{itemize}

This negative result reinforces our main finding: the purpose of warmup data is to anchor the critic to accurate values in the base policy's region, not to explore the state-action space.
The probe trajectories, despite their structured design, introduce transitions from perturbed behaviors that are irrelevant or misleading for the initial phase of value learning.
The simplest approach—collecting data from the base policy alone—remains the most effective.

\subsubsection{Q-Value Grounding Error}
\label{app:grounding_error}

To verify the value anchor effect quantitatively, we measure the \textit{Q-value grounding error}: the discrepancy between the critic's estimates and the true returns on base policy trajectories.

\paragraph{Definition.}
Let $\tau = (s_0, a_0, r_0, s_1, \ldots)$ be a trajectory collected by rolling out the base policy $\pi_{\text{base}}$.
The Monte Carlo return from state $s_t$ is:
\begin{equation}
G_t^{\pi_{\text{base}}} = \sum_{k=0}^{T-t} \gamma^k r_{t+k}
\end{equation}
The grounding error at timestep $t$ is:
\begin{equation}
\mathcal{E}_{\text{grounding}}(s_t, a_t) = \left| Q_\theta(s_t, a_t) - G_t^{\pi_{\text{base}}} \right|
\end{equation}
We report the average grounding error over a held-out set of base policy trajectories collected before training begins.

\paragraph{Justification I: Why Grounding Error over TD Error?}
A natural alternative would be to track the temporal difference (TD) error during training.
However, TD error measures \textit{Bellman consistency}, i.e., how well the critic aligns with its own bootstrapped target, rather than alignment with the true value landscape.
In the cold start phase, the target network $Q_{\theta'}$ is randomly initialized and unreliable.
Consequently, a low TD error may merely indicate that the critic has learned to predict the output of another random network (\textit{bootstrapping bias}), without learning the underlying task structure.
In contrast, the Monte Carlo return $G^{\pi_{\text{base}}}$ provides an \textit{unbiased} reference derived directly from environmental rewards, independent of any neural network estimates.
By measuring the grounding error against this external reference, we verify whether the critic is genuinely anchored to the physical value landscape rather than merely self-consistent.

\paragraph{Justification II: Why Base Policy Trajectories?}
Our metric evaluates the critic on pure base policy trajectories, while the critic ultimately learns to evaluate the combined policy $\pi_{\text{base}} + \lambda \cdot \pi_{\text{res}}$.
This approximation is justified in our setting for two reasons.
First, at the start of training, $\pi_{\text{res}}$ is initialized to output near-zero actions, making the combined policy nearly identical to $\pi_{\text{base}}$.
Second, accurate value estimates for $\pi_{\text{base}}$ are a prerequisite for guiding residual learning—the critic cannot improve upon a baseline it does not understand.
The grounding error thus measures exactly what matters for cold start: whether the critic has learned the value landscape in the region where residual learning will operate.

\paragraph{Results.}
As shown in Figure~\ref{fig:grounding} of the main text, the grounding error exhibits starkly different dynamics depending on whether warmup data is provided:
\begin{itemize}[leftmargin=*,itemsep=2pt,topsep=2pt]
    \item \textbf{Without warmup data}: The error briefly decreases in the first few thousand steps but then diverges rapidly. This pattern suggests that the critic initially makes progress but, lacking a stable anchor, eventually overfits to spurious patterns and drifts away from the true value landscape.
    
    \item \textbf{With warmup data}: The error drops rapidly at the start of training and remains low throughout. The warmup transitions provide a stable foundation that grounds the critic to the relevant region before residual learning begins.
\end{itemize}

This quantitative analysis confirms that warmup data serves as a value anchor: it grounds the critic to the true value landscape around $\pi_{\text{base}}$, enabling meaningful policy improvement from the start of training.
\subsection{Explicit Warmup}
\label{app:explicit_warmup}

This section provides additional details on the explicit warmup experiments discussed in Section~\ref{sec:explicit_warmup}.

\subsubsection{Experimental Protocol}
\label{app:explicit_protocol}

We investigate whether explicit critic pre-training can further accelerate learning beyond the implicit warmup provided by data anchoring.
During the explicit warmup phase, we freeze the residual policy $\pi_{\text{res}}$ and train only the critic for a fixed number of steps before joint actor-critic training begins.

\paragraph{Setup.}
We use 20K warmup transitions collected from the base policy (as established in Section~\ref{sec:warmup_anchor}).
The explicit warmup phase consists of 10K gradient steps on the critic, corresponding to half of the warmup data being sampled on average per update.
After the warmup phase, training proceeds normally with joint actor-critic updates.

\paragraph{Variants.}
We consider three representative variants that span the main design choices for critic pre-training:

\begin{itemize}[leftmargin=*,itemsep=6pt,topsep=4pt]
    \item \textbf{Explicit Soft Q (Auto $\alpha$)}: Standard SAC critic updates with automatic entropy tuning. The critic is updated by minimizing:
    \begin{equation}
    \mathcal{L}_Q = \mathbb{E}_{(s,a,r,s') \sim \mathcal{D}} \left[ \left( Q_\theta(s,a) - y \right)^2 \right]
    \end{equation}
    where the target $y$ includes the entropy term:
    \begin{equation}
    y = r + \gamma \left( \min_{i \in \{1,2\}} Q_{\theta'_i}(s', a') - \alpha \log \pi(a' \mid s') \right), \quad a' \sim \pi(\cdot \mid s')
    \end{equation}
    The entropy coefficient $\alpha$ is updated automatically to maintain a target entropy level:
    \begin{equation}
    \mathcal{L}_\alpha = \mathbb{E}_{a \sim \pi} \left[ -\alpha \left( \log \pi(a \mid s) + \bar{\mathcal{H}} \right) \right]
    \end{equation}
    where $\bar{\mathcal{H}}$ is the target entropy (typically $-\dim(\mathcal{A})$).
    
    \item \textbf{Explicit Soft Q (Fixed $\alpha$)}: Same as above, but the entropy coefficient is fixed at its initial value ($\alpha = 0.01$) throughout the warmup phase, disabling automatic tuning.
    The target remains:
    \begin{equation}
    y = r + \gamma \left( \min_{i \in \{1,2\}} Q_{\theta'_i}(s', a') - \alpha \log \pi(a' \mid s') \right)
    \end{equation}
    
    \item \textbf{Explicit Hard Q}: Critic updates using the standard Bellman target without entropy regularization:
    \begin{equation}
    y = r + \gamma \min_{i \in \{1,2\}} Q_{\theta'_i}(s', a'), \quad a' \sim \pi(\cdot \mid s')
    \end{equation}
    This removes the entropy term entirely during warmup, avoiding the entropy-related pathologies but introducing an objective mismatch with the subsequent Soft Q training.
\end{itemize}

In all variants, the residual policy $\pi_{\text{res}}$ remains frozen during the warmup phase, outputting near-zero actions with minimal variance due to its initialization.

\begin{figure}[b]
\centering
\includegraphics[width=0.4\linewidth]{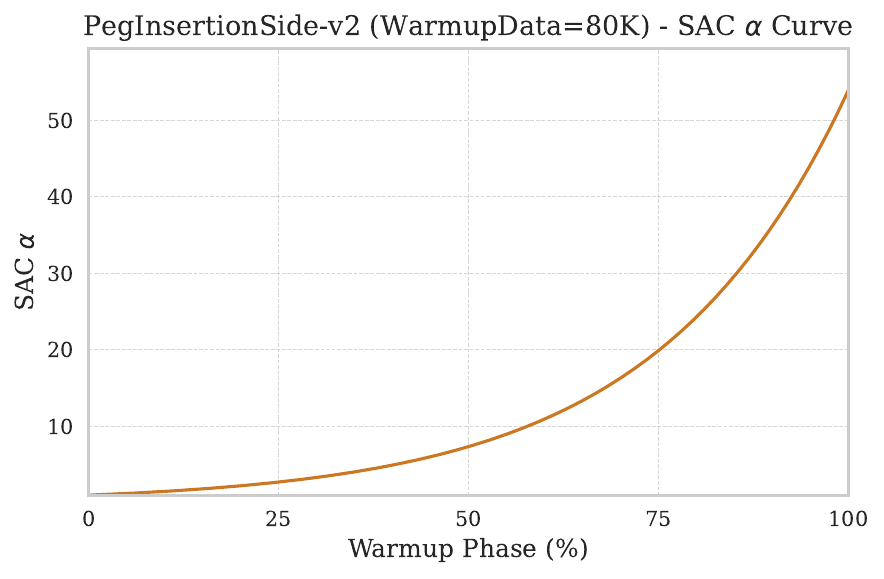}
\caption{\textbf{Effect of initial entropy coefficient on $\alpha$ divergence.}
Larger initial values lead to more severe divergence during explicit warmup.
With $\alpha_{\text{init}} = 1.0$, the entropy coefficient explodes rapidly, reaching values orders of magnitude higher than with $\alpha_{\text{init}} = 0.01$.}
\label{fig:alpha_init}
\end{figure}

\subsubsection{Effect of Initial Entropy Coefficient}
\label{app:alpha_init}

Section~\ref{sec:explicit_warmup} shows that with automatic entropy tuning, $\alpha$ diverges during the explicit warmup phase.
Here we examine how the initial value of $\alpha$ affects this divergence.

Figure~\ref{fig:alpha_init} compares the $\alpha$ dynamics during explicit warmup with different initial values.
The divergence is consistently observed across all initial values, but larger initial $\alpha$ leads to more severe and rapid divergence.
This is because the automatic tuning mechanism attempts to maintain a target entropy level.
With $\pi_{\text{res}}$ initialized to be near-deterministic, $|\log \pi|$ is extremely large, causing the tuning rule to continuously increase $\alpha$ in a futile attempt to reach the target entropy.
The larger the initial $\alpha$, the larger the entropy term $|\alpha \log \pi|$ becomes, which further destabilizes the TD target and accelerates the divergence.
This result confirms that the entropy dominance problem is fundamental to the explicit warmup setting and cannot be resolved by simply adjusting the initial entropy coefficient.

\subsubsection{Effect of Warmup Data Quantity}
\label{app:explicit_data_quantity}

One might hypothesize that explicit warmup fails because the warmup data (20K transitions) is insufficient for meaningful critic pre-training.
To test this, we increase the warmup data to 80K transitions and extend the explicit warmup phase to 40K gradient steps on the PegInsertionSide task.

\begin{figure}[ht]
\centering
\includegraphics[width=0.5\linewidth]{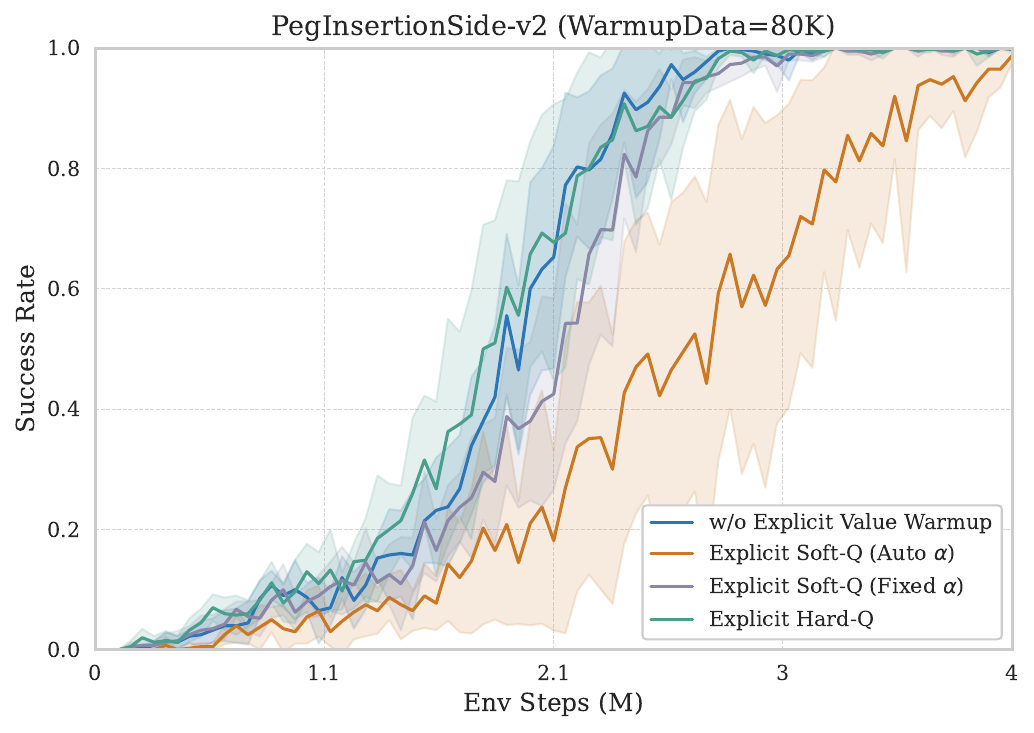}
\caption{\textbf{Increasing warmup data does not salvage explicit warmup.}
Even with 4$\times$ more warmup data (80K transitions) and proportionally longer warmup phase (40K steps), explicit Soft Q warmup still fails to improve over implicit warmup alone.}
\label{fig:explicit_80k}
\end{figure}

As shown in Figure~\ref{fig:explicit_80k}, increasing the quantity of warmup data does not resolve the failure of explicit warmup.
The entropy dominance problem persists regardless of data quantity: the near-deterministic initialization of $\pi_{\text{res}}$ makes $|\log \pi|$ extremely large, causing the entropy term to dominate the TD target and corrupt value learning.
This confirms that the failure of explicit warmup is not a matter of insufficient data, but a fundamental incompatibility between the SAC entropy mechanism and the residual policy initialization.
\section{Investigation of Scale Mismatch}

\subsection{Critic Normalization}
\label{app:normalization}

This section provides additional analysis on the normalization experiments in Section~\ref{sec:normalization}.

\subsubsection{The Necessity of Normalization}
\label{app:norm_necessity}

\begin{figure}[ht]
\centering
\includegraphics[width=0.85\linewidth]{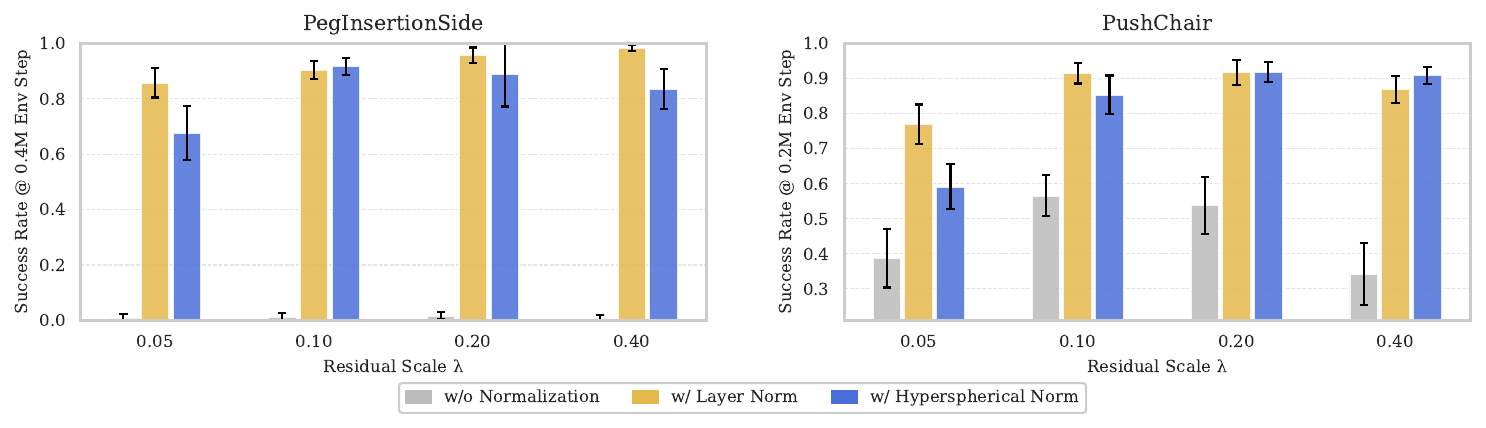}
\caption{\textbf{Normalization is essential across residual scales.}
Without normalization, the critic fails to learn efficiently regardless of $\lambda$.
Both LN and HN substantially outperform the unnormalized baseline, with the gap between LN and HN being much smaller than the gap between either and no normalization.}
\label{fig:norm_lambda}
\end{figure}

The main text compares Layer Normalization (LN) and Hyperspherical Normalization (HN).
Here we include the unnormalized baseline to emphasize that \textit{normalization itself} is the critical factor, while the choice between LN and HN is secondary.
Figure~\ref{fig:norm_lambda} shows performance across different residual scales $\lambda \in \{0.05, 0.1, 0.2, 0.4\}$.
The unnormalized critic fails almost entirely on PegInsertionSide regardless of $\lambda$, and performs substantially worse on PushChair.
In contrast, both LN and HN achieve strong performance across the range of $\lambda$ values.
The difference between LN and HN is minor compared to the dramatic improvement both provide over the unnormalized baseline.
This confirms that addressing the scale mismatch through normalization is the key insight; the specific normalization technique is a secondary design choice.

\subsubsection{Why Normalization Benefits the Critic but Not the Actor}
\label{app:actor_norm}

\begin{figure}[ht]
\centering
\includegraphics[width=0.7\linewidth]{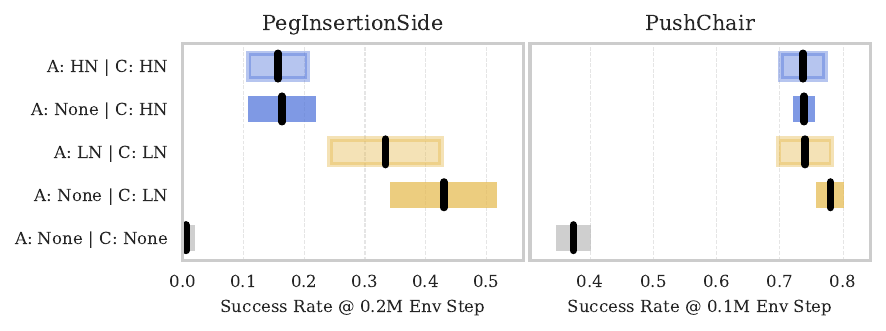}
\caption{\textbf{Normalization benefits the critic but not the actor.}
Adding normalization to the actor (A: LN or A: HN) provides no improvement over normalizing only the critic (A: None | C: LN or A: None | C: HN).
The bottom row shows that without critic normalization, performance collapses regardless of actor normalization.}
\label{fig:actor_critic}
\end{figure}

Figure~\ref{fig:actor_critic} presents a systematic ablation of normalization placement.
We compare five configurations: no normalization, critic-only normalization (with LN or HN), and full normalization of both actor and critic.

The results reveal a clear pattern:
\begin{itemize}[leftmargin=*,itemsep=2pt,topsep=2pt]
    \item \textbf{Critic normalization is necessary}: The bottom row (A: None | C: None) shows that without critic normalization, performance is severely degraded on both tasks.
    \item \textbf{Actor normalization is unnecessary}: Comparing "A: None | C: LN" with "A: LN | C: LN" shows that adding normalization to the actor provides no benefit and may slightly hurt performance on PegInsertionSide.
    \item \textbf{The pattern holds for both LN and HN}: The same trend appears with Hyperspherical Normalization.
\end{itemize}

This asymmetry confirms that scale mismatch is fundamentally a \textit{value learning} problem.
The critic takes the combined action $a = a_{\text{base}} + \lambda \cdot a_{\text{res}}$ as input, where the small $\lambda$ causes the residual component to be overshadowed by the base action.
Normalization decouples the critic's internal representations from input magnitudes, restoring its ability to detect residual variations.
The actor, in contrast, takes only the state as input and faces no such scale mismatch.
Once the critic provides meaningful gradients, the actor can learn effectively without additional architectural modifications.

\subsubsection{Diagnostic Metrics}
\label{app:diagnostic_metrics}

To understand \textit{how} normalization improves value learning, we introduce two diagnostic metrics visualized in Figure~\ref{fig:mechanism}.

\paragraph{Critic Sensitivity.}
We measure the critic's responsiveness to residual actions via the gradient norm:
\begin{equation}
\text{Sensitivity} = \mathbb{E}_{s \sim \mathcal{D}} \left[ \left\| \nabla_{a_{\text{res}}} Q_\theta(s, a_{\text{base}} + \lambda \cdot a_{\text{res}}) \right\| \right]
\end{equation}
This metric quantifies how strongly the critic's output responds to changes in the residual action.
A higher sensitivity indicates that the critic can detect small variations in $a_{\text{res}}$, which is essential for providing meaningful policy gradients.
Without normalization, the dominant magnitude of $a_{\text{base}}$ suppresses the critic's sensitivity to the residual component.

\paragraph{Value Difference.}
We measure the critic's ability to attribute value to the residual policy:
\begin{equation}
\text{Value Difference} = \mathbb{E}_{s \sim \mathcal{D}} \left[ \left| Q_\theta(s, a_{\text{base}} + \lambda \cdot a_{\text{res}}) - Q_\theta(s, a_{\text{base}}) \right| \right]
\end{equation}
This metric captures the value contribution that the critic assigns to the residual action.
As training progresses, a well-functioning critic should recognize that the residual improves upon the base policy and assign increasing value to it.
Without normalization, this growth is substantially slower, indicating the critic struggles to distinguish the residual's contribution.

\subsection{Distributional Objectives}
\label{app:distributional}

This section provides implementation details for the distributional RL experiments in Section~\ref{sec:distributional}.

\subsubsection{Implementation Details}
\label{app:distributional_impl}

We compare the standard MSE objective against three distributional alternatives, all implemented within the SAC framework.
The key modification for actor-critic integration is that the actor update uses the \textit{expected value} (mean of the distribution) rather than the full distribution, while the critic learns the complete return distribution.

\paragraph{C51.}
Categorical distributional RL~\citep{bellemare2017distributional} models the return distribution as a categorical distribution over a fixed set of atoms.
The critic outputs logits over $N$ atoms with fixed support $\{z_i\}_{i=1}^N$ uniformly spaced in $[V_{\min}, V_{\max}]$.
The distributional Bellman target is projected onto this support, and the critic is trained with cross-entropy loss.
For SAC integration, the entropy term is subtracted from the target values before projection:
\begin{equation}
\mathcal{T}Z(s,a) \stackrel{D}{=} r + \gamma \left( Z(s', a') - \alpha \log \pi(a' \mid s') \right)
\end{equation}
The actor is updated using the expected Q-value: $\mathbb{E}[Z(s,a)] = \sum_i p_i z_i$.

Our experiments use a shifted sparse reward $r_t = r_t^{\text{sparse}} - 1 \in \{-1, 0\}$.
For this setting, the theoretical value range is $V_{\min} = -(1 - \gamma^T)/(1 - \gamma)$ and $V_{\max} \approx 0$, where $T$ is the episode length.
Table~\ref{tab:c51_support} lists the support ranges used for each task.

\begin{table}[h]
\centering
\caption{C51 hyperparameters by task.}
\label{tab:c51_support}
\begin{tabular}{lcccc}
\toprule
Task & $\gamma$ & $T$ & $V_{\min}$ & $V_{\max}$ \\
\midrule
PegInsertionSide & 0.97 & 200 & $-35$ & $0$ \\
PushChair & 0.9 & 200 & $-10$ & $0$ \\
\bottomrule
\end{tabular}
\end{table}

\paragraph{QR-DQN.}
Quantile Regression~\citep{dabney2018distributional} learns a set of return quantiles without requiring a fixed support.
The critic outputs $N$ quantile estimates $\{\theta_i\}_{i=1}^N$ corresponding to quantile fractions $\tau_i = (2i-1)/(2N)$.
The loss is computed pairwise between all current quantiles and all target quantiles:
\begin{equation}
\mathcal{L}_{\text{QR}} = \frac{1}{N^2} \sum_{i=1}^{N} \sum_{j=1}^{N} \rho_{\tau_i}^\kappa \left( y_j - \theta_i(s, a) \right)
\end{equation}
where $y_j = r + \gamma \theta_j(s', a') - \alpha \log \pi(a' \mid s')$ is the $j$-th target quantile, and $\rho_\tau^\kappa$ is the quantile Huber loss that weights overestimation and underestimation asymmetrically based on $\tau$.
The actor uses the mean of the quantiles: $\bar{Q}(s,a) = \frac{1}{N}\sum_i \theta_i(s,a)$.

\begin{table}[ht]
\centering
\caption{QR-DQN hyperparameters.}
\label{tab:qr}
\begin{tabular}{lc}
\toprule
Hyperparameter & Value \\
\midrule
Number of quantiles $N$ & 25 \\
Huber threshold $\kappa$ & 1.0 \\
\bottomrule
\end{tabular}
\end{table}

\paragraph{TQC.}
Truncated Quantile Critics~\citep{kuznetsov2020controlling} extends quantile regression with an ensemble of critics and truncates the highest quantile estimates to mitigate overestimation bias.
When computing the target, we drop the top $d$ quantiles from each of the $K$ critics:
\begin{equation}
y = r + \gamma \left( \frac{1}{K(N-d)} \sum_{k=1}^{K} \sum_{i=1}^{N-d} \theta_{k,i}(s', a') - \alpha \log \pi(a' \mid s') \right)
\end{equation}
where $\theta_{k,i}$ denotes the $i$-th lowest quantile from the $k$-th critic.
This truncation provides a pessimistic value estimate that reduces the overestimation bias common in actor-critic methods.

\begin{table}[ht]
\centering
\caption{TQC hyperparameters.}
\label{tab:tqc}
\begin{tabular}{lc}
\toprule
Hyperparameter & Value \\
\midrule
Number of quantiles $N$ & 25 \\
Number of critics $K$ & 5 \\
Top quantiles dropped per critic $d$ & 2 \\
\bottomrule
\end{tabular}
\end{table}

\subsubsection{Value Distribution Analysis}
\label{app:anatomy}

Figure~\ref{fig:anatomy} in the main text visualizes the "anatomy" of residual value learning: despite the residual's negligible footprint in action space, its effect on Q-values manifests as a clear mean shift.
Here we describe the experimental design and provide additional analysis across training.

\paragraph{Setup.}
We sample 1024 states from the replay buffer and compute two Q-values for each state:
\begin{itemize}[leftmargin=*,itemsep=2pt,topsep=2pt]
    \item $Q(s, a_{\text{base}})$: the value of the base policy action alone.
    \item $Q(s, a_{\text{full}})$: the value of the combined action $a_{\text{base}} + \lambda \cdot a_{\text{res}}$.
\end{itemize}
We plot the distribution of these values across all sampled states.

\begin{figure}[ht]
\centering
\includegraphics[width=\linewidth]{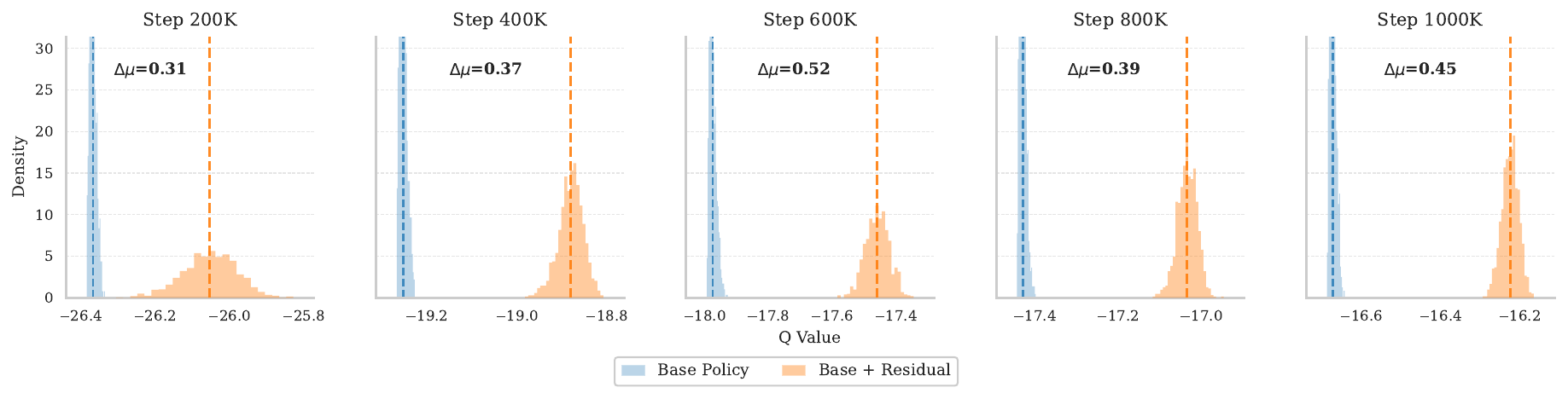}
\caption{\textbf{Q-value distributions throughout training.}
The mean shift between base policy (blue) and combined policy (orange) remains consistent across all training stages.
As training progresses, overall Q-values increase (from $\approx$-26 to $\approx$-16), reflecting improved policy performance, while the separation $\Delta\mu$ between the two distributions remains stable (ranging from 0.31 to 0.52).}
\label{fig:qvalue_evolution}
\end{figure}

\paragraph{Mean Shift Throughout Training.}
The main text shows the Q-value anatomy at a single checkpoint.
A natural question is whether this mean shift pattern persists throughout training or only appears at certain stages.
Figure~\ref{fig:qvalue_evolution} tracks the Q-value distributions at five checkpoints from 200K to 1000K environment steps on PegInsertionSide.

Two observations emerge from this analysis:
\begin{itemize}[leftmargin=*,itemsep=2pt,topsep=2pt]
    \item \textbf{Consistent separation}: The mean shift $\Delta\mu$ between base and combined policy Q-values remains stable throughout training, ranging from 0.31 to 0.52. This indicates that the critic consistently recognizes the value contribution of the residual policy at all stages of learning.
    \item \textbf{Shape preservation}: The distribution shapes remain similar across checkpoints; the primary change is in location (mean) rather than spread or modality. This reinforces our finding that the residual's effect on value is well-characterized by a simple mean shift.
\end{itemize}

These results provide additional evidence that MSE-based critics are well-suited for residual value learning: since the value signal consistently manifests as a mean shift rather than complex distributional changes, directly optimizing for mean differences is both sufficient and efficient.
\stopcontents[part2]

\resumecontents[part3]
\section{\DAWN~Implementation}
\label{app:implementation}

This section provides the complete algorithm and hyperparameters for \DAWN.

\subsection{Algorithm}
\label{app:algorithm}

\DAWN~consists of two key modifications to the standard residual RL pipeline: (1) implicit warmup via base policy data collection, and (2) critic normalization using Layer Normalization.
Algorithm~\ref{alg:dawn} presents the complete training procedure.

\begin{center}
\begin{minipage}{0.65\linewidth}
\begin{algorithm}[H]
\caption{\DAWN: Data Anchoring With Normalization}
\label{alg:dawn}
\begin{algorithmic}[1]
\REQUIRE Base policy $\pi_{\text{base}}$, residual scale $\lambda$, warmup transitions $N_{\text{warmup}}$
\STATE Initialize residual policy $\pi_{\text{res}}$ with near-zero output
\STATE Initialize critic $Q_\theta$ \hlc[cyan!20]{with Layer Normalization} \hfill \textcolor{cyan!70!black}{$\triangleleft$ \textit{Key modification}}
\STATE Initialize target critic $Q_{\theta'} \leftarrow Q_\theta$, replay buffer $\mathcal{D} \leftarrow \emptyset$
\STATE
\STATE \colorbox{orange!20}{\textcolor{gray}{\textit{// Phase 1: Value Anchoring}}} \hfill \textcolor{orange!70!black}{$\triangleleft$ \textit{Key modification}}
\WHILE{$|\mathcal{D}| < N_{\text{warmup}}$}
    \STATE $a_{\text{base}} \leftarrow \pi_{\text{base}}(s)$
    \STATE \hlc[orange!20]{Execute $a = a_{\text{base}}$, observe $r, s'$} \hfill \textcolor{gray}{\textit{// no residual, no training}}
    \STATE Store $(s, a_{\text{base}}, r, s')$ in $\mathcal{D}$
\ENDWHILE
\STATE
\STATE \textcolor{gray}{\textit{// Phase 2: Residual RL Training}}
\FOR{each environment step}
    \STATE $a_{\text{base}} \leftarrow \pi_{\text{base}}(s)$
    \STATE $a_{\text{res}} \sim \pi_{\text{res}}(\cdot \mid s)$
    \STATE Execute $a = a_{\text{base}} + \lambda \cdot a_{\text{res}}$, observe $r, s'$
    \STATE Store $(s, a_{\text{base}}, a_{\text{res}}, r, s')$ in $\mathcal{D}$
    \STATE
    \FOR{each gradient step}
        \STATE Sample mini-batch $\mathcal{B}$ from $\mathcal{D}$
        \STATE Update $Q_\theta$ by minimizing Bellman error (Eq.~\ref{eq:soft_bellman})
        \STATE Update $\pi_{\text{res}}$ by maximizing $\mathbb{E}[Q_\theta(s, a_{\text{base}} + \lambda \cdot a_{\text{res}}) - \alpha \log \pi_{\text{res}}]$
        \STATE Soft update: $Q_{\theta'} \leftarrow \tau Q_\theta + (1-\tau) Q_{\theta'}$
    \ENDFOR
\ENDFOR
\end{algorithmic}
\end{algorithm}
\end{minipage}
\end{center}

The key differences from prior residual RL methods are:
\begin{itemize}[leftmargin=*,itemsep=2pt,topsep=2pt]
    \item \textbf{Warmup data uses base policy only}: Unlike exploration-based warmup strategies, we collect transitions using $\pi_{\text{base}}$ alone with zero residual actions. This grounds the critic to the true value landscape around the base policy.
    \item \textbf{No progressive exploration}: We omit the $\epsilon$-schedule used in Policy Decorator~\citep{yuan2025policy}. With efficient value learning via normalization, this protective mechanism becomes unnecessary.
    \item \textbf{Critic normalization}: The critic uses Layer Normalization to address scale mismatch, enabling it to detect small residual contributions.
\end{itemize}

\subsection{Hyperparameters}
\label{app:hyperparameters}

To ensure fair comparison with Policy Decorator~\citep{yuan2025policy}, we adopt the same residual scale $\lambda$ for each task.
Table~\ref{tab:lambda_all} lists the complete hyperparameters across all experimental configurations.

\begin{table}[ht]
\centering
\caption{Residual scale $\lambda$ across all tasks and base policy configurations. Values are adopted from Policy Decorator for fair comparison.}
\label{tab:lambda_all}
\begin{tabular}{llc}
\toprule
Task & Base Policy & $\lambda$ \\
\midrule
\multicolumn{3}{l}{\textit{Diffusion Policy (state observations)}} \\
\quad PegInsertionSide & Diffusion Policy & 0.1 \\
\quad TurnFaucet & Diffusion Policy & 0.1 \\
\quad PushChair & Diffusion Policy & 0.2 \\
\quad Pen & Diffusion Policy & 0.2 \\
\quad Hammer & Diffusion Policy & 0.1 \\
\quad Relocate & Diffusion Policy & 0.1 \\
\midrule
\multicolumn{3}{l}{\textit{Diffusion Policy (visual observations)}} \\
\quad TurnFaucet & Diffusion Policy & 0.05 \\
\quad PushChair & Diffusion Policy & 0.2 \\
\midrule
\multicolumn{3}{l}{\textit{Behavior Transformer (state observations)}} \\
\quad Door & BeT & 0.3 \\
\quad Pen & BeT & 0.3 \\
\quad Hammer & BeT & 0.3 \\
\quad Relocate & BeT & 0.2 \\
\bottomrule
\end{tabular}
\end{table}

\DAWN~only introduces one additional hyperparameter: the number of warmup transitions $N_{\text{warmup}}$.
Based on our investigation in Section~\ref{sec:warmup_anchor}, we use $N_{\text{warmup}} = 20\text{K}$ across all tasks.
This value provides sufficient data for value anchoring while adding negligible overhead to the total training budget.

\paragraph{Integration with Base Policies.}

\DAWN~is agnostic to the base policy architecture.
Here we describe the integration details for the two base policy types used in our experiments.

\begin{itemize}[leftmargin=*,itemsep=2pt,topsep=2pt]
    \item \textbf{Diffusion Policy}: Diffusion Policy~\citep{chi2025diffusion} outputs a sequence of actions via iterative denoising. Following Policy Decorator, we use action chunking: the policy predicts 16 future actions, of which 4 are executed before re-planning. The residual policy outputs a corresponding sequence of 4 residual actions, which are added to the base actions element-wise after scaling by $\lambda$. The critic receives the summed action sequence flattened into a single vector.
    \item \textbf{Behavior Transformer}: BeT~\citep{shafiullah2022behavior} also uses action chunking with a context window. The integration follows the same pattern: the residual policy outputs actions matching the base policy's action horizon, and the critic evaluates the summed actions.
\end{itemize}

\section{Additional Ablations}
\label{app:ablation}

This section presents additional ablation studies to validate \DAWN's design choices.

\subsection{Sensitivity to Residual Scale}
\label{app:lambda_sensitivity}

\begin{figure}[ht]
\centering
\includegraphics[width=\linewidth]{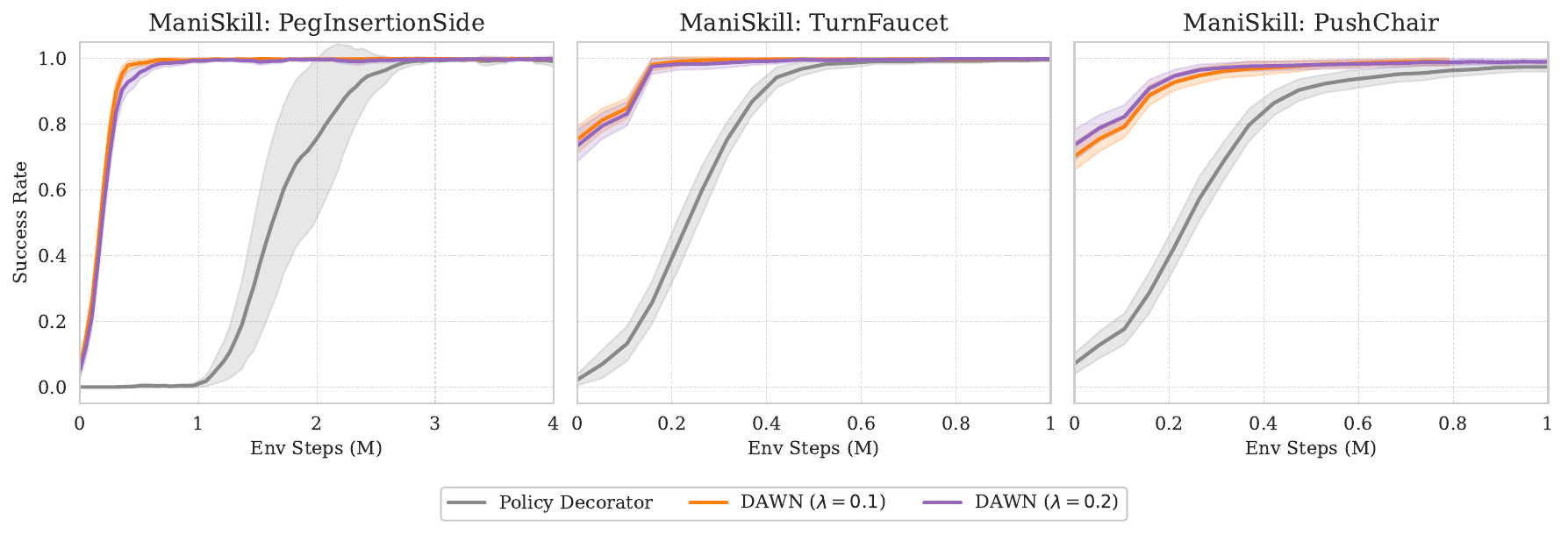}
\caption{\textbf{Sensitivity to residual scale $\lambda$.}
\DAWN~maintains strong performance across different $\lambda$ values (0.1 and 0.2), consistently outperforming Policy Decorator.
The minimal gap between \DAWN~($\lambda=0.1$) and \DAWN~($\lambda=0.2$) indicates consistent robustness.}
\label{fig:lambda_sensitivity}
\end{figure}

The residual scale $\lambda$ controls the magnitude of corrections the residual policy can apply.
A natural concern is whether \DAWN~is sensitive to this hyperparameter.
Figure~\ref{fig:lambda_sensitivity} evaluates \DAWN~with two different $\lambda$ values against Policy Decorator on three ManiSkill tasks.

The results demonstrate that \DAWN~is robust to the choice of $\lambda$:
\begin{itemize}[leftmargin=*,itemsep=2pt,topsep=2pt]
    \item On PegInsertionSide, both $\lambda$ values achieve superior efficiency, converging to 100\% success rate before 1M steps.
    \item On TurnFaucet and PushChair, the performance gap between $\lambda=0.1$ and $\lambda=0.2$ is negligible.
    \item Across all tasks, both \DAWN~variants substantially outperform Policy Decorator in sample efficiency.
\end{itemize}

This robustness simplifies hyperparameter tuning in practice: practitioners can select $\lambda$ based on task-specific considerations (e.g., desired correction magnitude) without concerns about value learning stability.

\subsection{Progressive Exploration is Unnecessary}
\label{app:prog_explore}

Policy Decorator~\citep{yuan2025policy} introduces progressive exploration, which gradually increases the probability of using the residual policy during training (Equation~\ref{eq:progressive}).
This mechanism was designed to stabilize training when value learning is inefficient.
We hypothesize that with efficient value learning via \DAWN, this protective mechanism becomes unnecessary.

\begin{figure}[ht]
\centering
\includegraphics[width=\linewidth]{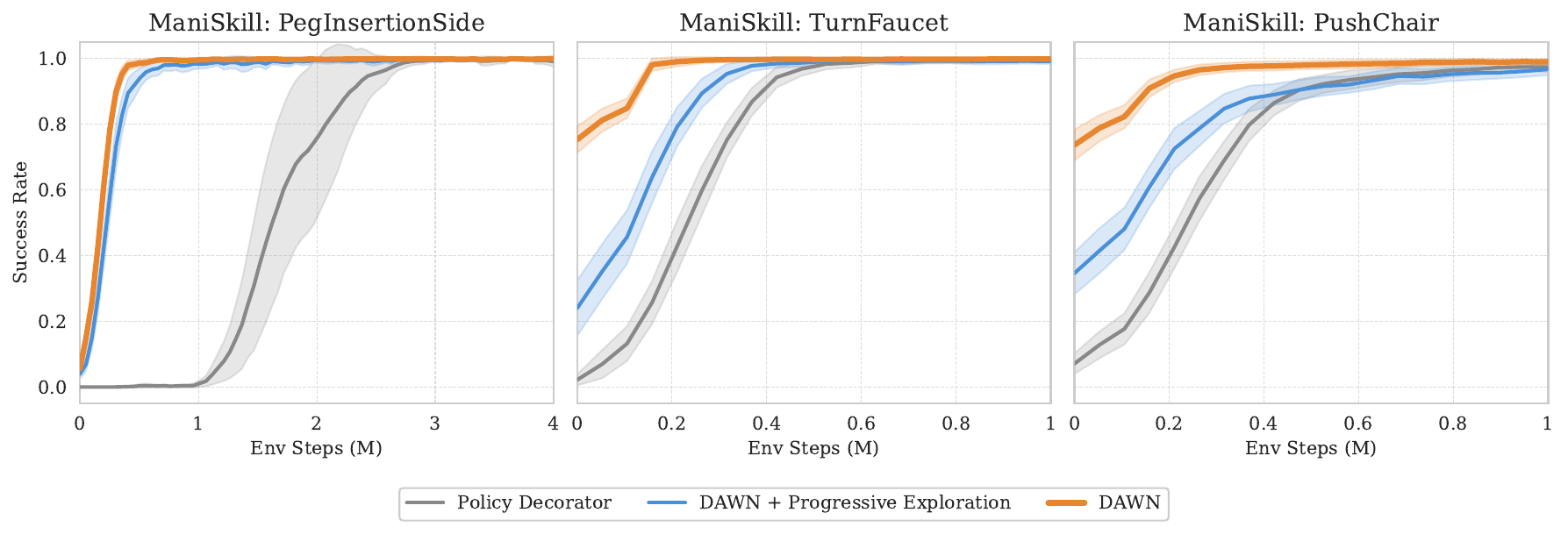}
\caption{\textbf{Progressive exploration is unnecessary with efficient value learning.}
\DAWN~without progressive exploration outperforms Policy Decorator even when the latter uses a carefully tuned progressive schedule.
This confirms that addressing the root causes of inefficient value learning eliminates the need for protective heuristics.}
\label{fig:prog_explore}
\end{figure}

Figure~\ref{fig:prog_explore} compares three configurations: (1) Policy Decorator without progressive exploration (our minimal baseline), (2) Policy Decorator with progressive exploration using the tuned schedule from~\citet{yuan2025policy}, and (3) \DAWN~without progressive exploration.
The progressive exploration schedules use $H = 30\text{K}$ for PegInsertionSide, $H = 100\text{K}$ for TurnFaucet, and $H = 300\text{K}$ for PushChair.

The results reveal a clear pattern:
\begin{itemize}[leftmargin=*,itemsep=2pt,topsep=2pt]
    \item \textbf{Progressive exploration helps the baseline}: Adding progressive exploration to Policy Decorator improves performance, particularly on PegInsertionSide. This confirms its role as a protective mechanism against unstable value learning.
    \item \textbf{\DAWN~outperforms both}: Without any progressive exploration, \DAWN~achieves better sample efficiency than Policy Decorator with progressive exploration. This demonstrates that addressing the root causes (cold start and scale mismatch) is more effective than symptom-level fixes.
    \item \textbf{Simpler and more efficient}: \DAWN~eliminates the need for task-specific tuning of the schedule parameter $H$, which varies by an order of magnitude across tasks (30K to 300K in these experiments).
\end{itemize}

This ablation validates our design philosophy: rather than adding protective mechanisms to mask inefficient value learning, \DAWN~directly addresses the underlying pathologies, resulting in a simpler and more effective method.
\stopcontents[part3]


\end{document}